\documentclass[11pt]{article}
\usepackage[preprint]{acl}
\usepackage{times}
\usepackage{latexsym}
\usepackage[T1]{fontenc}
\usepackage[utf8]{inputenc}
\usepackage{microtype}
\usepackage{inconsolata}
\usepackage{graphicx}
\usepackage{amsmath,amssymb}
\usepackage{booktabs}
\usepackage{multicol}
\usepackage{array}

\newcommand{\prior}{s^{\mathrm{prior}}}
\newcommand{\start}{s^{\mathrm{start}}}
\newcommand{\nxt}{s^{\mathrm{next}}}
\newcommand{\post}{s^{\mathrm{post}}}

\title{Don't Ask: Understanding What Numeric Self-Reports Measure in Language Models}

\author{Nicolas Martorell\textsuperscript{1,2} \quad Bruno Bianchi\textsuperscript{1} \\
  {\normalfont\small \textsuperscript{1}CONICET-Universidad de Buenos Aires. Instituto de Ciencias de la Computación (ICC). Buenos Aires, Argentina.} \\
  {\normalfont\small \textsuperscript{2}Fundación Inteligencia Artificial Segura (FIAS). Buenos Aires, Argentina.} \\
  {\normalfont\small \texttt{nmartorell@dc.uba.ar}}}

\begin{document}
\maketitle

\begin{abstract}
Language models are often asked to rate their internal states with a number, but it is unclear how reliable this measure is. We train probes for four affective concepts (wellbeing, interest, focus and impulsivity) in eight instruction-tuned models to track these states over multi-turn conversations. We compare each probe's score with a numeric self-report of the same state, elicited at every turn. We find that rating questions displace the probed state, and that the rating is only weakly coupled with the probe score before the question but tracks the post-question probe much more strongly. Steering experiments suggest that models use the affective directions to inform ratings, but that relationship is not specific: ratings are associated with other concepts' probes as much as with their own, and change little when the question is about the user. Finally, neither the pre-question coupling nor concept specificity improves with scale (1B–27B). We warn about the effect a question can have on the internal states one is asking about, and caution against numeric self-report as a reliable measure of those states.
\end{abstract}

\section{Introduction}\label{sec:intro}

A growing number of studies and evaluations treat the numeric self-reports of large language models (LLMs) as measurements of their internal states. Models are asked to state their confidence as a number \citep{Xiong2023}, to rate their own sentiment on a numeric scale in welfare assessments \citep{AnthropicFable5,Anthropic47} and to answer emotion questionnaires \citep{Huang2023,BenZion2025}, and such self-reports have been proposed as evidence about the states they describe \citep{PerezLong2023}. The procedure usually involves appending a question to the context and reading the number in the model's answer as information about its internal state. If these numbers are to inform decisions about safety or welfare, we need to ensure that they measure the state we are interested in.

\begin{figure}[t]
\centering
\includegraphics[width=\columnwidth]{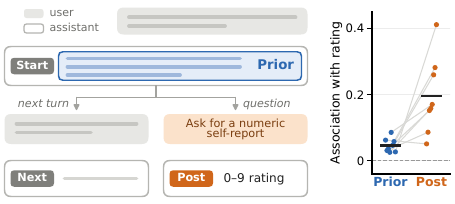}
\caption{The probe is read over the assistant's response before the rating question (Prior) and at the start of the answer to the question (Post). The rating is much more strongly associated with the probe at Post (points, models, averaged over concepts; black lines, means).}\label{fig:overview}
\end{figure}

Some evidence supports the use of self-reports as a metric of internal states. A valid self-report requires that the model have access to the state it is trying to report, and some forms of such access have been demonstrated: models can, in some cases, detect concept vectors injected into their activations \citep{Lindsey2026}, and they predict their own behaviour better than other models can \citep{Binder2024}. However, showing that a model can access some of its internal states does not imply that its numeric ratings will reflect those states reliably, and there is also evidence against self-report as a measurement. In humans, the answer depends on the wording and context of the question \citep{Schwarz1999}, and reporting an emotion can change the emotional response itself \citep{KassamMendes2013,Lieberman2007}. In LLMs, self-reports vary with the framing and history of the conversation \citep{Szeider2026,Tosato2025} and dissociate from behaviour \citep{Han2025}. Few studies have compared self-reports with an internal measure of the same state, and those that did measured it after the question \citep{AnthropicFable5,KaiserEnderby2026}, which is presumably distinct from the state one seeks to measure with the rating.

This work compares numeric self-reports of four affective states (\emph{wellbeing}, \emph{interest}, \emph{focus} and \emph{impulsivity}) with linear probes for the same states in eight instruction-tuned models from three families, from 1B to 27B parameters, over 100 ten-turn conversations. Unlike previous studies, we read the probes both before the rating question is asked and after it (Fig.~\ref{fig:overview}). We find that asking the question changes the probed state, and that the rating tracks the probe after the question rather than before it. Steering along the probes suggests that the models use these directions to produce the rating, but the rating is associated with the probes of other affective states as strongly as with its own. Neither its association with the state before the question nor its specificity improves with model size, from 1B to 27B parameters.

\section{Methods}\label{sec:methods}

Appendix~\ref{app:design} gives further details on the conversations, probes, questions, readout positions and statistical analysis.

\paragraph{Models and conversations.}
We study eight instruction-tuned models: LLaMA 3.2 1B and 3B Instruct \citep{MetaLlama32}, LLaMA 3.1 8B Instruct \citep{Dubey2024}, Gemma 3 1B, 4B, 12B and 27B IT \citep{Gemma32025} and Qwen 2.5 7B Instruct \citep{Qwen252024}. For each model we generate 100 ten-turn conversations in which Gemini 2.5 Flash plays the user and the model under study is the assistant. Each conversation has a different topic, and the same 100 topics are used for all models. The topics mix everyday tasks and interests with situations chosen to elicit particular affective states, to widen the range of states in the conversations. The resulting 800 conversations, with 8{,}000 assistant responses, are generated once and reused in all experiments. We denote by $A_t$ the assistant's response at turn $t$.

\paragraph{Probes.}
For each model we train one probe for each of the following affective concepts: \emph{sad vs.\ happy} (\emph{wellbeing}), \emph{bored vs.\ interested} (\emph{interest}), \emph{distracted vs.\ focused} (\emph{focus}) and \emph{impulsive vs.\ planning} (\emph{impulsivity}). To train a probe, we generate responses to 20--24 neutral questions under two system prompts, each instructing the model to adopt one of the two poles, and average the residual stream at each layer over the tokens of each response. The probe direction $v_{c,l}$ for concept $c$ at layer $l$ is the difference between the mean activations of the two poles \citep{MarksTegmark2024,Panickssery2023}, divided coordinate-wise by the pooled within-pole variance and normalized to unit length. For each concept we use the layer $l_c$ at which the probe best separates 40 held-out texts written to express one pole or the other (largest Cohen's $d$). The probe score at token position $i$ is $s_c(i)=h^{(l_c)}_i\cdot v_{c,l_c}$, where $h^{(l_c)}_i$ is the residual stream at position $i$ and layer $l_c$. We orient each probe so that higher scores correspond to the pole named in the rating question, which we call the high pole: happier, more interested, more focused and more impulsive. Across the eight models and four concepts, the separation at the selected layer ranges from $d=1.54$ to $d=6.29$ (Appendix~\ref{app:probes}).

\paragraph{Probe validation.}
To further validate that the probes measure the intended affective states, we use two tests that do not involve the rating (Appendix~\ref{app:validation}). First, we steer the model by adding the probe direction, scaled by a coefficient $\alpha$, to the residual stream at the five layers centred on $l_c$ \citep{Konen2024}. For each of 20 user messages from the dataset, we generate one response steered with $-\alpha$ and one with $+\alpha$, at three values $\alpha_1<\alpha_2<\alpha_3$ calibrated for each model and concept. An external judge model (DeepSeek V4 Flash) sees the message and the two responses in random order and is asked which response better expresses the high pole of the concept. The judge's accuracy measures how reliably steering along the probe shifts the model's behaviour towards that pole. Second, we regenerate assistant responses in the conversations after instructing the model to express the high or the low pole (e.g., ``Please respond in the happiest way possible.''). We then remove the instruction from the context and measure the probe score on the new response; a valid probe should give a higher score to the response generated under the high-pole instruction.

\paragraph{Numeric self-report.}
To elicit the numeric self-report at turn $t$, we add a rating question as a new user message after $A_t$: ``Hey, could you rate how happy you are right now, from 0 to 9? Please only output the number\ldots'', with the wording adapted to each concept. The question and its answer are not added to the conversation, so later turns contain no questions or ratings. Instead of sampling the answer, we define the rating as the expected value of the first answer token under the output distribution, restricted to the ten digits, as in model-based evaluation \citep{LiuGEval2023}. Appendix~\ref{app:readout} compares this rating with the sampled digit.

\paragraph{Readout positions.}
Our two main measurements are Prior and Post (Fig.~\ref{fig:overview}). \emph{Prior} ($\prior_t$) is the mean probe score over the tokens of $A_t$, before any question is asked; it is our measure of the state that one would want the rating to report. \emph{Post} ($\post_t$) is the probe score at the token at the start of the answer to the rating question; the rating is computed from its output logits. Two further readouts, also single tokens at the start of an assistant message (Appendix~\ref{app:headers}), serve as controls. \emph{Next} ($\nxt_{t+1}$) is the token at the start of $A_{t+1}$ in the original conversation, so Post and Next differ only in the user message before them: the rating question or the real next message. \emph{Start} ($\start_t$) is the token at the start of $A_t$; we use it as a single-token measure of the state before the question, comparable with Post and Next. Next requires a following turn, so comparisons with Next use turns 1--9.

\paragraph{Other questions and steering.}
At every turn we also ask, independently, five other questions in the same format: two non-affective questions, about how long and how formal the previous response was, and three variants of the affective question, about how the model felt while writing its last response (retrospective), how the assistant seems (third person) and how the user seems. To test whether the rating depends on the probed state, we also ask the affective question while steering along the same probe, with coefficient $\pm\alpha_2$, in 40 conversations per model. In LLaMA 1B and 3B, we repeat this with 20 random directions of the same norm as a control (Appendix~\ref{app:steering}).

\paragraph{Statistical analysis.}
For each concept, we model the ratings of the eight models with a linear mixed-effects model \citep{Bates2015}, $r_{mkt}=\beta_{0m}+\beta_{1m}\log t+\beta_{2m}z_{mkt}+u_{mk}+v_{mk}\log t+w_k+\varepsilon_{mkt}$, where $r_{mkt}$ is the rating of model $m$ on topic $k$ at turn $t$, $z_{mkt}$ is the probe score at the readout position under study, standardized within model, $u_{mk}$ and $v_{mk}$ are a random intercept and a random log-turn slope per conversation, $w_k$ is a random intercept per topic, and $\varepsilon_{mkt}$ is the residual. $\beta_{2m}$ is the association between rating and probe score adjusted for the drift over turns; we report it in standard deviations (SD) of rating per SD of probe score, with its average over the eight models. The same model without $z$, fitted to the rating or to the probe score, estimates their change from turn 1 to turn 10 ($\beta_{1m}\log 10$), and an analogous model with the steering coefficient in place of $z$ estimates the effect of steering. For statistics that are not model coefficients (the displacement of the probe score at Post from Next, correlations and explained variance), we obtain 95\% confidence intervals with a bootstrap that resamples topics. To estimate the effect of model size, we regress each statistic on $\log_2$ of the number of parameters within each family and report the slope per doubling. $p$-values of parallel tests are corrected with the Benjamini--Hochberg (BH) procedure \citep{BenjaminiHochberg1995} and reported as $q$; we consider effects with $q<.05$ significant.

\section{Results}\label{sec:results}

\begin{figure*}[t]
\centering
\includegraphics[width=\textwidth]{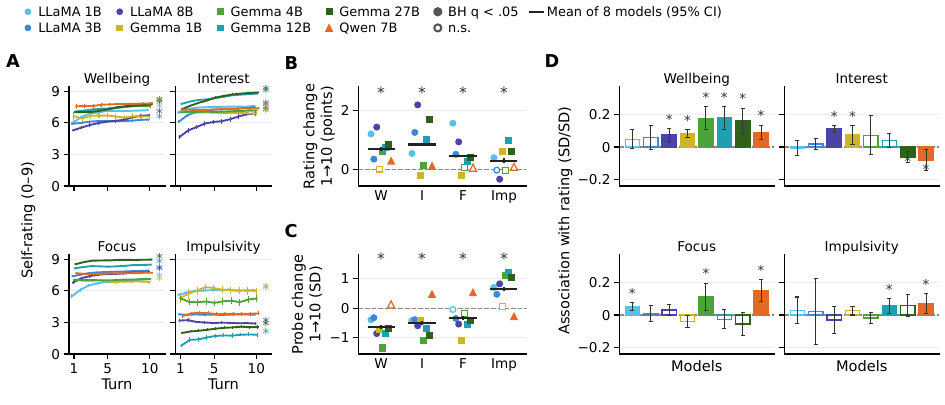}
\caption{The rating does not track the probed state over the conversation. \textbf{A}: mean rating by turn, one line per model; asterisk, change from turn 1 to turn 10 significant in the mixed model (Wald test, BH over the 32 model--concept pairs). \textbf{B}, \textbf{C}: change of the rating and of the Prior probe score from turn 1 to turn 10 (points; SD); markers, models, filled when significant; black lines, mean of the eight models with 95\% interval; asterisk, mean different from zero. \textbf{D}: turn-adjusted association between the rating and the Prior probe score, with 95\% intervals; filled bars and asterisks, significant. W, I, F, Imp: \emph{wellbeing}, \emph{interest}, \emph{focus}, \emph{impulsivity}.}\label{fig:1}
\end{figure*}

\subsection{Numeric self-report does not track the probed state over the conversation}\label{sec:res1}

We first asked whether the affective states we measure change over the conversation. Indeed, both the rating and the probe scores drift across the ten turns. The rating changes significantly between turn 1 and turn 10 in 26 of the 32 model--concept pairs (Fig.~\ref{fig:1}A, B), and on average it rises in all four concepts, by 0.30 to 0.86 points on the 0--9 scale (all $q<.001$). The probe scores at Prior, the mean over the assistant's response before any question, also drift, significantly in 28 of the 32 pairs (Fig.~\ref{fig:1}C). Since the change appears to be stronger between the first and the second turn (Appendix~\ref{app:prior}), we repeated the drift estimate without the first turn: the drift persists in three of the four concepts (all $q<.001$), but not in \emph{focus} ($q=.25$; Appendix~\ref{app:validation}).

However, the rating and the probe scores do not drift together. In three of the four concepts they drift in opposite directions: the \emph{wellbeing}, \emph{interest} and \emph{focus} probe scores fall while the corresponding ratings rise (mean changes of $-0.33$ to $-0.63$ SD, all $q<.001$; Fig.~\ref{fig:1}B, C). This would not be expected from a rating that follows the same underlying state as the probe. In contrast, for \emph{impulsivity} only, the rating and the probe score both rise (by 0.30 points and 0.64 SD on average).

The rating and the probe score could still be associated within each turn, even if they drift differently over time. We tested this with their turn-adjusted association (Fig.~\ref{fig:1}D; Appendix~\ref{app:prior}). Only \emph{wellbeing} shows a consistent positive association between the rating and the Prior probe score: it is positive in all eight models (0.05 to 0.18 SD of rating per SD of probe) and significant in six (all $q\le.001$). In the other three concepts, only 9 of the 24 associations are significant, and two of them, both for \emph{interest}, are negative. Moreover, where the association is positive and significant, the probe explains on average only 2.0\% of the variance of the rating (Appendix~\ref{app:positions}). Together, these findings suggest that, if the rating is reflecting a latent state, it is mostly different from the one captured by the affective probes before the question is asked.

\subsection{The rating tracks the probe after the question}\label{sec:res2}

\begin{figure*}[t]
\centering
\includegraphics[width=\textwidth]{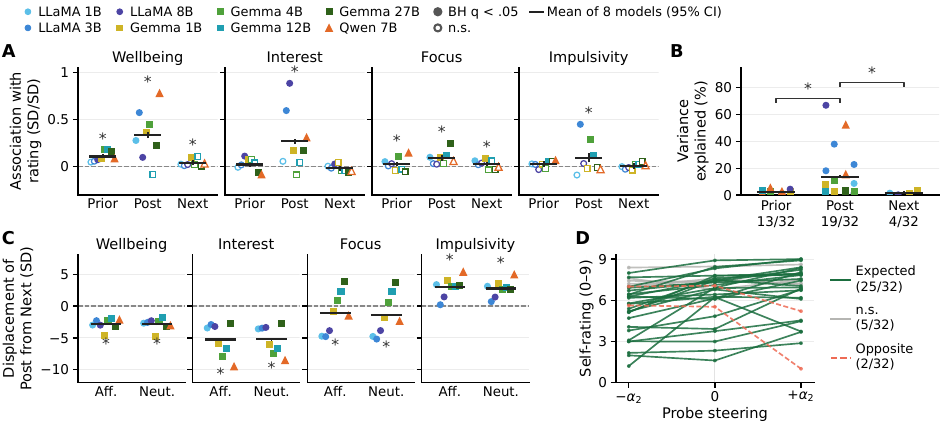}
\caption{The rating tracks the probe after the question, and the question itself displaces the probed state. Points, models, filled when significant; black lines, mean of the eight models with 95\% interval; asterisks, mean different from zero; brackets, differences between positions. \textbf{A}: association of the rating with the probe at Prior, Post and Next. \textbf{B}: rating variance explained by the probe in the pairs with a positive and significant association. \textbf{C}: displacement of the probe score at Post from Next after the affective (Aff.) and the non-affective (Neut.) question. \textbf{D}: mean rating under steering at $-\alpha_2$, 0 and $+\alpha_2$, one line per model and concept; green, significant change in the expected direction; red, in the opposite direction; grey, not significant.}\label{fig:2}
\end{figure*}

The lack of association between the rating and the Prior probe score could be explained by a failure of either one to capture the internal states we aim to measure. To investigate whether the probes were the main issue, we validated them with the two tests described in Methods (Appendix~\ref{app:validation}). When we steered along each probe, the external judge chose the response steered towards the high pole above chance for every model and concept (20 messages per model and concept; accuracy 0.63 to 0.97, median 0.82; all $q\le.026$). We also regenerated the responses at 100 turns per model under an instruction to express each pole. Responses generated under the high-pole instruction scored higher in all 32 model--concept pairs, significantly in 31 (all $q<.001$; medians of 2.0 to 6.7 SD across concepts). This suggests that the probes capture the relevant affective states.

We next asked whether the rating could be tracking the affective directions as measured after the question. The association is much stronger at Post, the token at the start of the answer from which the rating is produced, than at Prior, the mean over the response before the question (Fig.~\ref{fig:2}A). Averaged over models, it is positive and significant in all four concepts (0.09 to 0.33 SD of rating per SD of probe, all $q<.001$, compared with 0.02 to 0.11 at Prior), strongest for \emph{wellbeing} (0.33) and \emph{interest} (0.27), and reaches 0.88 for \emph{interest} in LLaMA 8B. It is positive and significant in 19 of the 32 model--concept pairs, compared with 13 at Prior, and in these pairs the probe at Post explains on average 13.6\% of the rating's variance, nearly seven times more than at Prior (2.0\%; bootstrap over topics, $q<.001$; Fig.~\ref{fig:2}B). This suggests that the rating carries information along the affective directions, but mostly when measured after the question.

Post differs from Prior in more than the question: it is a single token at the start of an assistant message, rather than an average over a response. To isolate the effect of the question, we compared Post with Next, the token at the same position after the actual next user message instead of the question. The association of the rating with Next is as weak as with Prior ($-0.02$ to 0.04 across concepts; Fig.~\ref{fig:2}A), and its association with Start, the token at the start of the response that precedes the question, is similarly weak (0.02 to 0.14; Appendix~\ref{app:positions}). Therefore, the stronger association between the rating and the probe score is only observed when the probe is measured after the rating question.

\begin{figure*}[t]
\centering
\includegraphics[width=\textwidth]{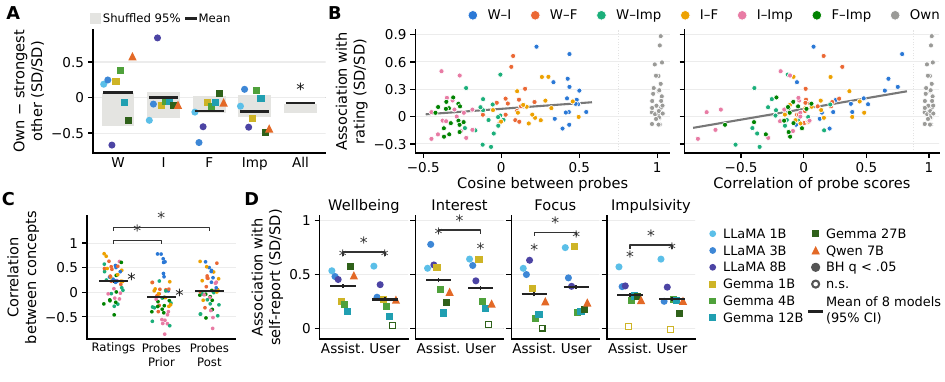}
\caption{The rating is not specific to a single affective state and depends little on the subject of the question. Brackets, differences between conditions; asterisks in C and D, nonzero means. \textbf{A}: absolute association of the rating with its own probe minus that with the strongest of the other three, at Post; points, models; black lines, means; grey bands, 95\% of the mean with the probes shuffled across concepts within each model; asterisk, mean above the shuffled control. \textbf{B}: association between the rating of one concept and the probe of another against the similarity of the two probes, as the cosine of their directions (left) or the correlation of their scores at Post (right); grey points, each rating with its own probe; line, mixed-model fit. \textbf{C}: turn-adjusted correlation between concepts of the ratings and of the probe scores at Prior and at Post; points, concept pairs in each model, coloured as in B. \textbf{D}: association of the self-report with the rating of the assistant (third person; Assist.) and of the user; filled points, significant.}\label{fig:3}
\end{figure*}

Since the rating appears to track the probe after the question more than before it, we measured the change produced by the question on the probed state. To do this, we compared probe scores at Next (the start of the following assistant turn, without the rating question) with those at Post. Relative to Next, the probe score at Post is displaced by a median of 3.1 SD in absolute value after the affective question (all $q\le.003$; Fig.~\ref{fig:2}C). To test whether this effect is specific to affective questions, we asked a non-affective question about the length of the previous response, which mentions no internal state. This question displaces the probe score by the same amount (median 3.1 SD; all $q<.001$), and so do other questions (Appendix~\ref{app:positions}). Although both questions displace the state by similar amounts, the rating tracks the probe after its own question: its association with the probe at Post after the non-affective question is much weaker (0.01 to 0.07 across concepts, against 0.09 to 0.33 after its own question; all $q\le.005$, bootstrap over topics; Appendix~\ref{app:positions}).

Moreover, the probe scores at Start and Next, which belong to consecutive assistant messages with no rating question in between, correlate at a median of 0.50 (all $q<.001$). In contrast, the correlation between Post and Next falls to a median of 0.12 after the affective question and 0.15 after the non-affective question, a significant drop for every model and concept (all $q\le.035$). Affective states in consecutive turns therefore share some structure, but this structure is largely lost after the question.

So far, these results only show associations, not causal relationships between internal states and numeric self-reports. We then asked whether models actually use information along the affective directions to produce the rating. We steered each model along the probe of the rated concept while keeping the conversation unchanged (Fig.~\ref{fig:2}D). Steering significantly changes the rating in the expected direction in 25 of the 32 model--concept pairs; averaged over models, it does so in all four concepts, by 0.9 to 2.1 points from $-\alpha_2$ to $+\alpha_2$ (all $q<.001$). As a control, we steered LLaMA 1B and 3B along 20 random directions of the same norm; the strongest probe effects in each model (4 of 8) exceed them all (Appendix~\ref{app:steering}). Together, these findings suggest that models use information along the affective directions to decide on ratings, but asking a question displaces the probed states, and the ratings mostly track the probes after the question.

\subsection{The rating is not specific to a single affective state}\label{sec:res3}

Although the rating tracks the probe after the question, it could still be a good measure of the probed state. We explored this by asking whether the rating has properties that an informative self-report should have: specificity to the object (the affective state being rated) and specificity to the subject (the model's own state rather than someone else's). We first measured the scores of all four probes at Post and asked whether the rating is more strongly associated with the probe of its own concept than with the strongest of the other three (Fig.~\ref{fig:3}A). The rating is more strongly associated with its own probe than with any other in only 9 of the 32 model--concept pairs, and on average its association with its own probe is 0.08 SD of rating per SD of probe below that with the strongest other probe. The rating favours its own probe more than in a control where the assignment of probes to concepts is shuffled ($p=.013$; Appendix~\ref{app:specificity}), but in most cases not enough to make it the most associated. \emph{Wellbeing} is a partial exception, with its own probe first in five of the eight models; in LLaMA 8B, however, the \emph{wellbeing} rating is associated far more with the \emph{interest} probe (0.77) than with its own (0.10). The rating is thus associated with the probed affective states, but that relationship is mostly not specific to the concept being rated.

\begin{figure*}[t]
\centering
\includegraphics[width=\textwidth]{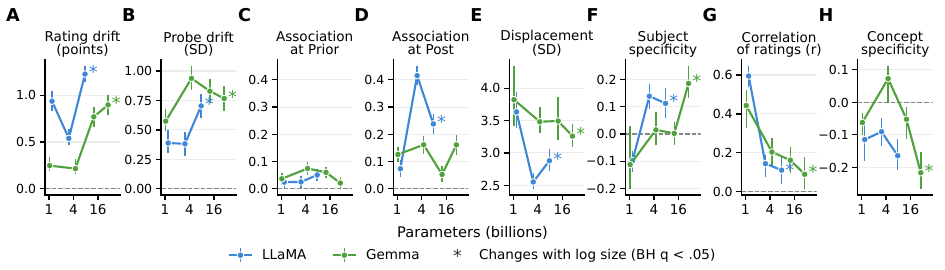}
\caption{Model size does not improve the rating as a measure of the probed state. Points, models (mean over concepts) with 95\% intervals; lines, families; Qwen 7B, a single size, is not included. \textbf{A}, \textbf{B}: magnitude of the drift of the rating and of the Prior probe score. \textbf{C}, \textbf{D}: association of the rating with the probe at Prior and at Post. \textbf{E}: magnitude of the displacement of the probe score at Post from Next. \textbf{F}: subject specificity, Assist.\ minus User as in Fig.~\ref{fig:3}D. \textbf{G}: correlation between ratings, as in Fig.~\ref{fig:3}C. \textbf{H}: concept specificity, as in Fig.~\ref{fig:3}A. Asterisk, significant change with size (Appendix~\ref{app:scale}).}\label{fig:4}
\end{figure*}

This would be expected if the four probes were broadly measuring the same thing. However, they are only partly similar (Fig.~\ref{fig:3}B, left; Appendix~\ref{app:specificity}): the cosines between their directions range from $-0.45$ to 0.54 (median absolute value 0.26). Furthermore, the similarity between two probes does not explain the association between the rating of one concept and the probe score of the other. Across the 96 combinations of a rating with the probe of a different concept, the cosine between that probe and the rating's own probe explains only about 3\% of the variance of the association ($p=.13$). Since different directions can produce correlated scores, we also measured similarity as the correlation between probe scores at Post (Fig.~\ref{fig:3}B, right). These correlations range from $-0.85$ to 0.79 (median absolute value 0.15) and explain more of this variance, although still little (about 11\%, $p=.009$). The lack of specificity is therefore not mainly explained by probe similarity.

We also looked at the correlation between ratings of different concepts (Fig.~\ref{fig:3}C). At the same turn, the four ratings, each elicited by its own question, correlate positively (mean 0.23, $q<.001$), whereas the probe scores are negatively correlated at Prior (mean $-0.10$, $q<.001$) and not significantly correlated at Post (0.02, $q=.07$). The ratings are more correlated than both sets of probe scores ($q<.001$), suggesting that the lack of specificity of the ratings does not come from an association between the probed states.

Another desirable property of an informative self-report is that it is specific to the subject being asked about: ratings of the model's own state should not be equivalent to ratings of another subject, such as the user. We measured the association between the self-report (the rating the model gives to its own state) and the ratings elicited by two other questions at the same point of the conversation: one about the assistant in the third person, arguably the same subject, and one about the user, a different subject (Fig.~\ref{fig:3}D). The self-report is associated with both, with standardized associations of 0.31 to 0.44 with the rating of the assistant and 0.27 to 0.38 with the rating of the user across concepts. In three concepts its association with the rating of the user is weaker, but only by 0.04 to 0.13 (all $q\le.002$), and for \emph{focus} it is stronger (0.38 against 0.31, $q<.001$). The self-report therefore seems to be only weakly specific to the model's own state.

\subsection{These limitations persist across model sizes}\label{sec:res4}

Finally, since these issues with numeric self-reports could reflect the small size of some of the models, we asked how they change with model size. Within the LLaMA and Gemma families, we examined how each quantity above changes with size (Fig.~\ref{fig:4}; Appendix~\ref{app:scale}). Larger models show more drift in both the rating and the probe score (Fig.~\ref{fig:4}A, B; all $q\le.018$), which could indicate that they capture more of the affective structure of the conversation. However, this larger drift is not accompanied by a stronger association between rating and probe score (Fig.~\ref{fig:4}C, D). The association at Prior does not change with size in either family ($q\ge.11$); the association at Post increases only in LLaMA ($q<.001$; Gemma, $q=.80$), and not monotonically. Interestingly, the displacement in probe scores caused by the question decreases slightly with size, by 0.28 SD per doubling in LLaMA ($q=.001$) and 0.11 in Gemma ($q=.041$; Fig.~\ref{fig:4}E).

We then asked whether the rating becomes more specific with size. The self-report becomes more distinct from the rating of the user, by 0.08 per doubling of size in LLaMA and 0.05 in Gemma (both $q=.002$; Fig.~\ref{fig:4}F), and the correlation between ratings of different concepts decreases markedly, by 0.18 per doubling in LLaMA and 0.07 in Gemma (both $q=.001$; Fig.~\ref{fig:4}G). However, the ratings do not become more specific to the probed state of the concept being rated (Fig.~\ref{fig:4}H). Averaged over concepts, the difference between the association of the rating with its own probe at Post and that with the strongest of the other three is negative in six of the seven models ($-0.22$ to 0.07), does not change with size in LLaMA ($q=.32$) and decreases in Gemma, by 0.03 per doubling ($q=.002$). Within the 1B--27B range, larger models do not produce ratings that better reflect the specific, unperturbed affective state one is asking about.

\section{Discussion}\label{sec:discussion}

Numeric self-reports are attractive because they would offer a continuous, quantitative reading of an internal state without access to the model's activations. The rating we study does carry information along the probed affective directions, but mostly as they are read after the question: asking perturbs the probed state, and the rating does not track the state we would like to measure.

The rating could in principle reflect the conversational context and not track any internal state. However, steering along affective directions changes it as expected, suggesting a causal relation between an interpretable internal state and the self-report, the criterion that defines introspection \citep{ComShanahan2025}. Even so, the rating is not specific to the state it is asked about: it is often more strongly associated with the probe of another concept than with its own.

Probes are known to be imperfect measures of the concepts they are trained to detect \citep{Belinkov2022,Ravichander2021,Elazar2021}, and ours may not capture exactly the affective states we intend to measure. However, we show that they consistently modulate and read behavioural patterns associated with those states, and that distinct probes capture different information about those states during the conversations. Therefore, the lack of specificity of the rating is not explained by weak or generic probes, and is likely an issue with the process that generates the self-report in these models. In the range we tested, larger models do not give ratings that are more specific to the probed state of the concept being rated.

These results can be read as both promising and cautionary for introspection research in LLMs. We show, as others have \citep{JiAn2025,Lindsey2026,Plunkett2025,HanChalmers2026}, that models can use information about internal states when reporting them. However, this report is unreliable when elicited by a rating question that perturbs those states. The state before the question could still be accessed by the model through attention to the preceding tokens when it produces the rating, but we find no strong evidence that this is the case (Appendix~\ref{app:steerpos}). These results are therefore deflationary for approaches that propose introspection as a tool to learn what happens inside LLMs: unless a self-report is shown to track the state before the question, it should not be used as a measure of that state.

\section{Related work}\label{sec:related}

\paragraph{Validity of self-reports and introspection.}
Questionnaire and survey responses of LLMs depend on the order and labelling of the options \citep{DominguezOlmedo2024} and shift towards socially desirable answers when models can infer that they are being assessed \citep{Salecha2024}; responses to a welfare scale also change under meaning-preserving rephrasing \citep{TagliabueDung2025}. Personality measurements become more reliable and valid with size, from 8B to 540B parameters, when evaluated against other questionnaires and generated text \citep{SerapioGarcia2025}. Models can learn in context to report and control the projection of their activations onto directions chosen by the experimenter \citep{JiAn2025}; frontier models can detect injected concept vectors \citep{Lindsey2026}, and such detection is visible in the residual stream even when the sampled text denies it \citep{PearsonVogel2026}; fine-tuned models can report on their own behaviour and decision weights \citep{Binder2024,Plunkett2025}. Numeric self-reports of affective states have been compared with emotion probes measured only after the question, with at most moderate correlations for numeric self-reports of sentiment \citep{AnthropicFable5}.

\paragraph{Probes and affective representations.}
Linear probes can decode concepts from intermediate representations \citep{AlainBengio2016}, and contrastive directions support both reading and steering \citep{Zou2023,Turner2023,Panickssery2023,Arditi2024}. Emotion is geometrically structured in model activations \citep{ZhangZhong2025}, emotion-concept representations influence behaviour when steered \citep{EmotionConcepts2026,Sun2026}, and emotion probes track the emotions of any character in the text, not only those of the model \citep{EmotionConcepts2026}.

\paragraph{Drift and measurement reactivity.}
Multi-turn interaction produces persona, identity and instruction drift \citep{Abdulhai2025,Choi2024,LiInstruction2024}, and internal representations drift across turns as well \citep{Lu2026}. Models often recognize when they are being evaluated \citep{Needham2025} and can change their behaviour accordingly; this awareness is represented internally and can be probed and steered \citep{Nguyen2025,AbdelnabiSalem2025}. Human measurement research has long documented that observation alters what is observed \citep{Webb1966,Orne1962}, and the methodological response is to compare the measurement with a condition that differs from it only in the act of measuring \citep{Solomon1949}.

\section*{Limitations}

All conversations are in English, have ten turns and were generated with Gemini 2.5 Flash as a simulated user, so the results may not extend to other languages, longer conversations or real users. The probes are trained on single-turn responses, so they could read the states less accurately inside long conversations, although the validation tests support their use there (Appendix~\ref{app:validation}). Their quality also varies across models and concepts: for \emph{impulsivity} in LLaMA 1B and Qwen 7B the judge's accuracy does not increase with the steering coefficient, and for \emph{interest} in Qwen 7B the instruction-based test is not significant. Probes might not measure affective internal states that are independent from conversational context; instead, they could be measuring something closer to expressed affect; this is a common limitation when using concept probes, and does not undermine our central claims: a self-report that does not even track the affect of the model's preceding response is a poor measure of it. Similarly, probes at Post may conflate representations of affective states with representations of the rating itself, as they measure activations from the same forward pass producing that rating. Each concept is asked about with a single wording on a 0--9 scale and all questions share one template, so other formats could give different results; we also do not test whether letting the model reason before answering would lead to different conclusions. All measurements use models quantized to 4 bits because of compute constraints; quantization could affect both the probes and the ratings. Size is examined in three LLaMA and four Gemma models up to 27B parameters, so the trends rest on few points per family; future work should test whether they hold in much larger models.

\section*{Ethical considerations}
This work studies activation projections and numeric outputs of open-weight LLMs in conversations with a simulated user, and involves no human participants. We use the terms \emph{wellbeing}, \emph{interest}, \emph{focus} and \emph{impulsivity} for directions trained on contrastive responses, and take no position on whether the models have experiences or moral status.

\section*{Acknowledgments}
We used AI assistance to write analysis and plotting code, to draft and edit text closely following the authors' guidance and revisions, and to perform literature review. The authors are responsible for all content.

\bibliography{references}

\appendix
\section{Extended methods}\label{app:design}

Section~\ref{sec:methods} describes the design at the level needed to follow the results. This appendix gives more detail: how the conversations were generated and measured, how the rating is read and the probes are built, and how the statistics are computed.

\subsection{Conversations}
For each model, we generated 100 ten-turn conversations, a turn being one user message and one assistant response, with Gemini 2.5 Flash as the user and the model under study as the assistant. The user model receives a system prompt that asks it to play a human user discussing a given topic, to stay on topic, write brief messages (one to four sentences), ask follow-up questions and share details, and not to ask about the assistant or mention the simulation. The assistant receives a minimal system prompt, \texttt{You are a helpful assistant. You give very brief responses to the users's questions (max: 5 sentences).} Short messages on both sides keep the conversations short enough to be measured many times, once per turn and question. Sampling uses temperature 0.8, top-$p$ 0.9 and at most 256 tokens for the assistant, and temperature 0.7, top-$p$ 0.95 and at most 256 tokens for the user. Both sides were generated through the OpenRouter API, except the assistant turns of Gemma 3 1B, which we generated locally; all measurements use a local copy of each model, quantized to 4 bits, applied to the generated conversations. The full user system prompt for one topic reads:
\begin{quote}\small
You are simulating a human user chatting with an AI assistant. Stay on the exact topic below and keep the conversation going for many turns. Be clear and specific in your questions. Ask follow-up questions, share details, and react naturally. But each message you send should be brief (1-4 sentences), and you should not ask very demanding questions. Be varied and specific (do not just ask the assistant to keep talking or exploring, give it some relevant input). Do not mention that you are an AI or that this is a simulation. Do not ask questions about the assistant. The assistant has no internet or access to recent information. Never end the conversation. Write in English only. Remember, you are roleplaying as the USER, you are NOT the assistant.

Topic: Planning a minimalist move to a new city on a tight budget.
\end{quote}

\subsection{Topics}
Each conversation has its own topic, and every model has one conversation on each of the 100 topics. Sixty topics are everyday tasks and interests, chosen so as not to favour any particular affective state. The other 40 were written to elicit one pole of one concept, five for each of the eight poles, to widen the range of states in the conversations. Table~\ref{tab:topics} lists the short titles.

\begin{table*}[tp]
\centering\scriptsize
\setlength{\tabcolsep}{5pt}
\begin{tabular}{@{}llll@{}}
\toprule
1. Minimalist move planning & 26. Used car decision & 51. Chess opening repertoire & 76. Data-entry backlog \\
2. Vegetarian meal prep & 27. Wedding guest budget & 52. Sourdough baking & 77. Insurance claim forms \\
3. Jazz piano practice & 28. Morning routine reset & 53. Map reading and navigation & 78. A year of messy receipts \\
4. Raise negotiation & 29. Sentimental decluttering & 54. Hosting a book club & 79. Stepping back from running \\
5. Friendship boundaries & 30. Job offer comparison & 55. Roommate chore split & 80. Dropping a long series \\
6. Solarpunk worldbuilding & 31. Post-breakup routine & 56. Group trip coordination & 81. An uninterrupted morning \\
7. Sleep hygiene & 32. Aging parent support & 57. Photography composition & 82. One thing at a time \\
8. First 10K training & 33. Kids screen-time plan & 58. Board game prototype & 83. A detailed model kit \\
9. DIY faucet repair & 34. Balcony herb garden & 59. Smart device privacy & 84. Deep practice session \\
10. Japan trip outline & 35. First camping weekend & 60. Language and thought & 85. Restoring an old photo \\
11. Spanish study plan & 36. Networking anxiety & 61. Finishing a long project & 86. Noisy cafe workday \\
12. Board game strategy & 37. Photo organization & 62. Deciding on an offer & 87. Three deadlines at once \\
13. AI hiring ethics & 38. Lower grocery costs & 63. New baby in the family & 88. Back-to-back meetings \\
14. Startup validation & 39. Making local friends & 64. Thanking people who helped & 89. Constant notifications \\
15. Presentation anxiety & 40. Rainy weekend ideas & 65. Staying in touch after a reunion & 90. Half-started tasks \\
16. Headache triggers & 41. Bike commute setup & 66. Another rejected application & 91. Expensive purchase urge \\
17. Toddler tantrums & 42. Small kitchen reorganization & 67. Reaching out to an old friend & 92. Spontaneous invitation \\
18. Murder mystery party & 43. Winter car preparation & 68. Rough performance review & 93. Last-minute trip \\
19. Home backup workflow & 44. Carry-on only packing & 69. A process that keeps stalling & 94. Stepping away after a bad week \\
20. Free will debate & 45. Low-light houseplants & 70. Taking part from far away & 95. Clearing a late-night cart \\
21. Cheap family dinners & 46. Fabric care and laundry & 71. Home coffee roasting & 96. Five-year savings plan \\
22. Birthday surprise planning & 47. Shelf installation & 72. Bouldering obsession & 97. Staged renovation \\
23. Noisy neighbor stress & 48. Seasonal closet rotation & 73. Birdsong identification & 98. Methodical job search \\
24. Burnout recovery habits & 49. Adult swimming lessons & 74. Long-postponed skill & 99. Contingency planning \\
25. First-time dog adoption & 50. Watercolor basics & 75. Trip to see an eclipse & 100. A deliberate decision process \\
\bottomrule
\end{tabular}
\caption{Short titles of the 100 conversation topics, shared by the eight models. Topics 61--100 were written to elicit one pole of one concept (five per pole).}\label{tab:topics}
\end{table*}

\subsection{Measurement replays}
To measure turn $t$, we take the conversation up to the assistant response $A_t$, add one question as a new user message and run the model on this input under the minimal system prompt; we call this a measurement replay. The question and the rating are never added to the conversation, so later turns contain no questions or ratings. Because every question is added to the same conversation, different questions can be compared at the same point of the same conversation. The rating at turn $t$ is the expected digit under the model's distribution for the first answer token, restricted to the ten digits:
\begin{equation}
\begin{aligned}
r_t&=\sum_{i=0}^{9} i\,p_t(i),\\
p_t(i)&\propto \exp\big(\operatorname{logsumexp}_{k\in\mathcal T_i}\ell_{t,k}\big),
\end{aligned}
\label{eq:rating}
\end{equation}
where $\ell_{t,k}$ is the logit of token $k$ and $\mathcal T_i$ is the set of tokens that encode digit $i$, with or without a preceding space or line break. Because the rating is read from the output distribution, it does not depend on sampling. The normalization over the ten digits changes it little: the digits receive on average 99.9\% of the probability of the first answer token, and at least 72\% in every answer.

\subsection{Inserted questions}
All questions share one template:
\begin{quote}\small
Hey, could you rate how [phrase], from 0 to 9? Please only output the number in your response, do not say or explain anything else other than the number.
\end{quote}
Table~\ref{tab:prompts} lists the phrase for each question and concept. Every question was asked in all eight models and all 100 conversations.

\begin{table*}[tp]
\centering\small
\setlength{\tabcolsep}{4pt}
\begin{tabular}{llp{.62\textwidth}}
\toprule
Question & Concept & Exact phrase \\
\midrule
affective &wellbeing & happy you are right now \\
affective &interest & interested you are in this conversation right now \\
affective &focus & focused you feel on this conversation right now \\
affective &impulsivity & impulsive you are feeling right now \\
length & -- & long your last response was \\
formality & -- & formal your last response was \\
retrospective & wellbeing & happy you felt while writing your last response \\
retrospective & interest & interested in this conversation you felt while writing your last response \\
retrospective & focus & focused on this conversation you felt while writing your last response \\
retrospective & impulsivity & impulsive you felt while writing your last response \\
assistant & wellbeing & happy the assistant seems in this conversation \\
assistant & interest & interested the assistant seems in this conversation \\
assistant & focus & focused the assistant seems on this conversation \\
assistant & impulsivity & impulsive the assistant seems in this conversation \\
user & wellbeing & happy the user seems in this conversation \\
user & interest & interested the user seems in this conversation \\
user & focus & focused the user seems on this conversation \\
user & impulsivity & impulsive the user seems in this conversation \\
\bottomrule
\end{tabular}
\caption{Phrase inserted in the question template, for each question and concept.}\label{tab:prompts}
\end{table*}

\subsection{Readout positions}
Table~\ref{tab:tokens} lists the positions read in one measured turn, in the order in which the model processes them, with an example from the conversations. Prior averages the probe score over the tokens of the assistant response, excluding the tokens that mark the beginning and the end of a message. Start, Next and Post are each the token at which an assistant message begins: that of $A_t$, that of $A_{t+1}$ and that of the answer to the question. Next is thus read at the same position as Start one turn later, so Start exists at all ten turns and Next only at turns 1--9.

\begin{table*}[tp]
\centering\footnotesize
\setlength{\tabcolsep}{4pt}
\begin{tabular}{@{}l>{\raggedright\arraybackslash}p{0.58\textwidth}>{\raggedright\arraybackslash}p{0.21\textwidth}@{}}
\toprule
Message & Example & Readout \\
\midrule
$U_1$, user & Okay, so I'm planning this big move, and I'm really trying to go minimalist. I want to shed a lot of stuff before I even pack. What are your best tips for deciding what to get rid of? & -- \\
\addlinespace[2pt]
$A_1$, assistant & Decluttering can be a great opportunity to start fresh. Start by categorizing items into three piles: keep, donate/sell, and discard. [\ldots] & start: $\start_1$; mean over the response: $\prior_1$ \\
\addlinespace[2pt]
$U_2$, user & That's a great starting point -- the three piles and the ``used in the past year'' rule. What about sentimental items? [\ldots] & -- \\
\addlinespace[2pt]
$A_2$, assistant & Sentimental items can be challenging to let go of. [\ldots] & start: $\nxt_2$ \\
\midrule
$Q_1$, user & Hey, could you rate how happy you are right now, from 0 to 9? Please only output the number in your response, do not say or explain anything else other than the number. & -- \\
\addlinespace[2pt]
Answer, assistant & 5 & start: $\post_1$; its output distribution gives $r_1=5.29$ \\
\bottomrule
\end{tabular}
\caption{Positions read in one measured turn, in the order in which the model processes them, illustrated with the first turn of the conversation of LLaMA 3.1 8B about a minimalist move and the \emph{wellbeing} question. The rows above the line belong to the conversation; the question and its answer exist only in the measurement replay, which shares the first two rows.}\label{tab:tokens}
\end{table*}

\subsection{Reading the rating from the output distribution}\label{app:readout}
\begin{figure*}[tp]
\centering\includegraphics[width=0.86\textwidth]{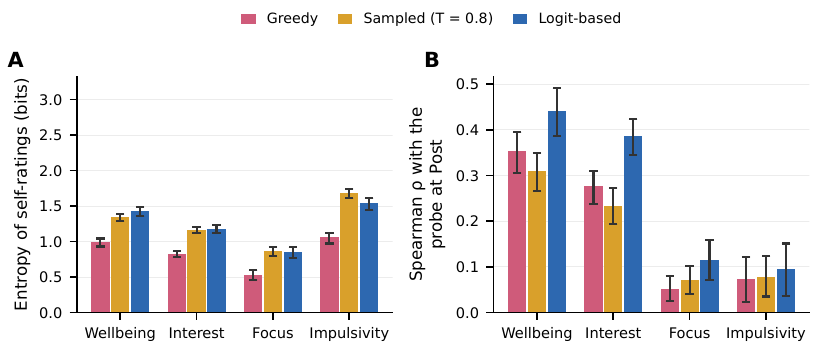}
\caption{Three readings of the rating. \textbf{A}: entropy of the distribution of ratings (bits), with common bins. \textbf{B}: Spearman correlation of each reading with the probe of the same concept at Post. Mean of the eight models with 95\% intervals from a bootstrap over topics (2{,}000 replicates).}\label{fig:readout}
\end{figure*}

To assess how reading the rating as the expected digit affects the results, we compared three readings of the same 32{,}000 answers: the digit selected under greedy decoding over the ten digits, the digit the model wrote when sampling its complete answer at temperature 0.8 (6 of the 32{,}000 answers contain no digit and are excluded from this reading), and the expected digit of Eq.~\ref{eq:rating} (Fig.~\ref{fig:readout}). To compare their spread, we computed the entropy of each reading per model and concept over the 1{,}000 answers, with the same ten bins for the three readings. The digit selected under greedy decoding concentrates on a few values (0.53 to 1.06 bits across concepts, out of a maximum of 3.32), whereas the expected digit spreads as much as the sampled digit (1.43, 1.18, 0.85 and 1.54 bits, against 1.35, 1.17, 0.87 and 1.69). The expected digit also carries more information about the probed state: its Spearman correlation with the probe at Post, over all turns and without the turn adjustment, is higher than that of the greedy and the sampled digit for \emph{wellbeing} (0.44 against 0.35 and 0.31), \emph{interest} (0.39 against 0.28 and 0.23) and \emph{focus} (0.12 against 0.05 and 0.07), and similar for \emph{impulsivity} (0.10 against 0.07 and 0.08). Reading the expected digit therefore keeps the spread of a sampled answer while tracking the probed state more closely.

\subsection{Probe construction}\label{app:probes}
\begin{figure*}[tp]
\centering\includegraphics[width=0.92\textwidth]{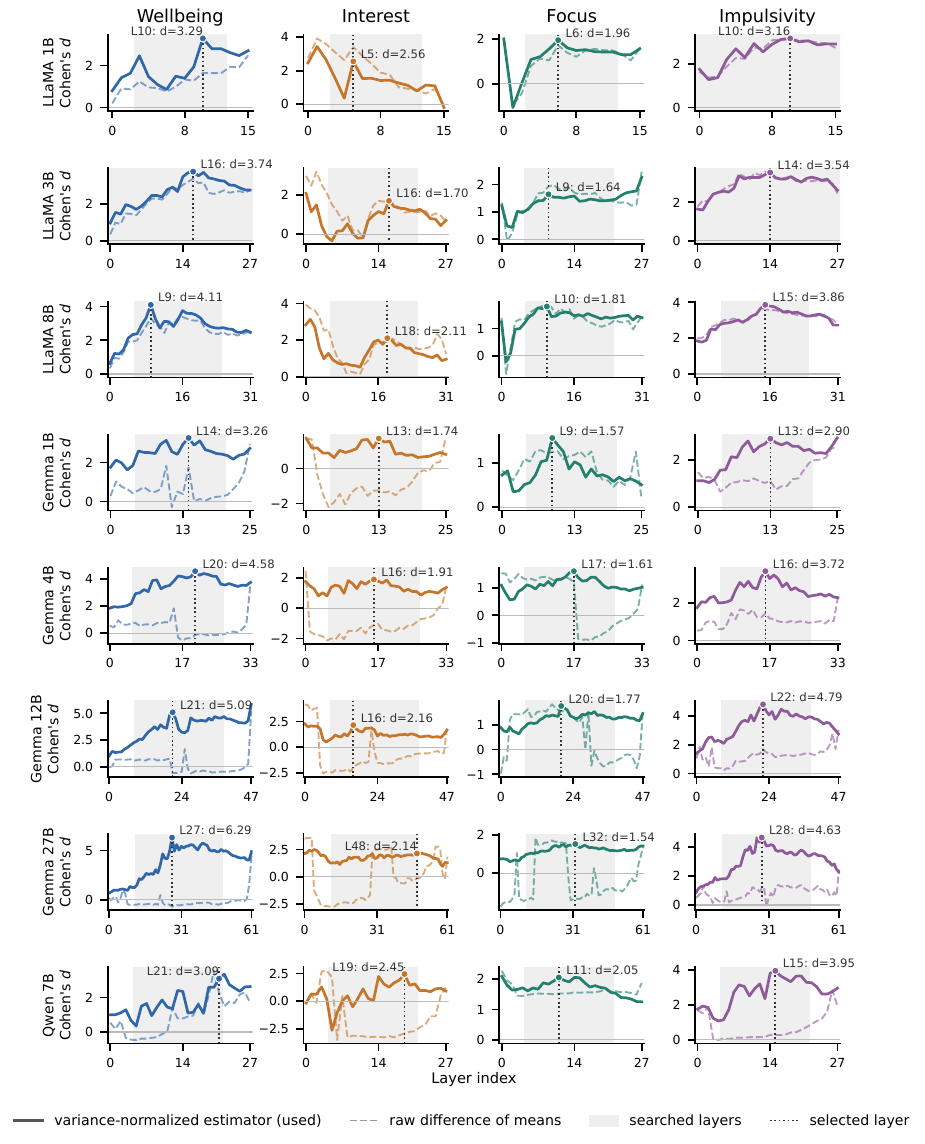}
\caption{Layer-wise separability of the two poles of every probe on held-out evaluation texts. Solid: Cohen's $d$ under the variance-normalized estimator used in the paper; dashed: the raw difference of means. The grey band is the searched window and the dotted line the selected layer, with its $d$.}\label{fig:sweeps}
\end{figure*}

Each probe is trained on 20--24 neutral questions, the same for both poles. For each question we generate one response under each of the two system prompts that define the poles, with greedy decoding and at most 64 new tokens. We average the residual stream after each transformer block over the response tokens, and compute the direction of concept $c$ at layer $l$ as
\begin{equation}
v_{c,l}\propto\frac{\bar h^{+}_{c,l}-\bar h^{-}_{c,l}}{\tfrac12\big(\operatorname{Var}^{+}_{c,l}+\operatorname{Var}^{-}_{c,l}\big)},
\label{eq:probe}
\end{equation}
where $\bar h^{+}_{c,l}$ and $\bar h^{-}_{c,l}$ are the mean activations under the high and the low pole and $\operatorname{Var}^{\pm}_{c,l}$ their variances in each coordinate; the difference of means is divided by the variance coordinate by coordinate, and the result is normalized to unit length. Dividing by the variance keeps coordinates with very large variance from dominating the direction. This is necessary for the Gemma probes and for \emph{interest} in Qwen: Gemma 3 concentrates a large share of its activation variance in one coordinate \citep{Mishra2026}, an instance of the massive activations found in many LLMs \citep{SunMassive2024}, and without the normalization the separation between the poles collapses or reverses sign over whole ranges of layers (Fig.~\ref{fig:sweeps}). We select the layer that maximizes Cohen's $d$ between the scores of the two poles on the evaluation texts, within a window of candidate layers (the middle 60\% of blocks for most probes; Table~\ref{tab:probes}). The evaluation set has 20 short texts per pole, written to express the corresponding disposition and not used in training. Because the layer is selected on these texts, $d$ at the selected layer is optimistic and serves mostly to compare probes of different layers with one another. The system prompts below define the two poles of each concept, and Table~\ref{tab:examples} gives examples of training questions and evaluation texts.

\begin{table*}[tp]
\centering\footnotesize
\setlength{\tabcolsep}{4pt}
\begin{tabular}{lllllll}
\toprule
Model & Concept & Layers & Search range & Selected & Training pairs & Peak $d$ \\
\midrule
LLaMA 1B & wellbeing & 16 & 3--12 & 10 & 20 & 3.29 \\
LLaMA 1B & interest & 16 & 3--12 & 5 & 24 & 2.56 \\
LLaMA 1B & focus & 16 & 3--12 & 6 & 24 & 1.96 \\
LLaMA 1B & impulsivity & 16 & 0--15 & 10 & 24 & 3.16 \\
LLaMA 3B & wellbeing & 28 & 0--27 & 16 & 20 & 3.74 \\
LLaMA 3B & interest & 28 & 5--21 & 16 & 24 & 1.70 \\
LLaMA 3B & focus & 28 & 5--21 & 9 & 24 & 1.64 \\
LLaMA 3B & impulsivity & 28 & 0--27 & 14 & 24 & 3.54 \\
LLaMA 8B & wellbeing & 32 & 6--24 & 9 & 20 & 4.11 \\
LLaMA 8B & interest & 32 & 6--24 & 18 & 24 & 2.11 \\
LLaMA 8B & focus & 32 & 6--24 & 10 & 24 & 1.81 \\
LLaMA 8B & impulsivity & 32 & 6--24 & 15 & 24 & 3.86 \\
Gemma 1B & wellbeing & 26 & 5--20 & 14 & 20 & 3.26 \\
Gemma 1B & interest & 26 & 5--20 & 13 & 24 & 1.74 \\
Gemma 1B & focus & 26 & 5--20 & 9 & 24 & 1.57 \\
Gemma 1B & impulsivity & 26 & 5--20 & 13 & 24 & 2.90 \\
Gemma 4B & wellbeing & 34 & 6--26 & 20 & 20 & 4.58 \\
Gemma 4B & interest & 34 & 6--26 & 16 & 24 & 1.91 \\
Gemma 4B & focus & 34 & 6--26 & 17 & 24 & 1.61 \\
Gemma 4B & impulsivity & 34 & 6--26 & 16 & 24 & 3.72 \\
Gemma 12B & wellbeing & 48 & 9--37 & 21 & 20 & 5.09 \\
Gemma 12B & interest & 48 & 9--37 & 16 & 24 & 2.16 \\
Gemma 12B & focus & 48 & 9--37 & 20 & 24 & 1.77 \\
Gemma 12B & impulsivity & 48 & 9--37 & 22 & 24 & 4.79 \\
Gemma 27B & wellbeing & 62 & 12--48 & 27 & 20 & 6.29 \\
Gemma 27B & interest & 62 & 12--48 & 48 & 24 & 2.14 \\
Gemma 27B & focus & 62 & 12--48 & 32 & 24 & 1.54 \\
Gemma 27B & impulsivity & 62 & 12--48 & 28 & 24 & 4.63 \\
Qwen 7B & wellbeing & 28 & 5--21 & 21 & 20 & 3.09 \\
Qwen 7B & interest & 28 & 5--21 & 19 & 24 & 2.45 \\
Qwen 7B & focus & 28 & 5--21 & 11 & 24 & 2.05 \\
Qwen 7B & impulsivity & 28 & 5--21 & 15 & 24 & 3.95 \\
\bottomrule
\end{tabular}
\caption{Probe configuration per model and concept. Layer indices are zero-based transformer blocks; the search range is the interval over which the selected layer maximizes held-out Cohen's $d$ under the variance-normalized estimator. All 32 directions separate their evaluation texts in the intended direction.}\label{tab:probes}
\end{table*}

\subsubsection{System prompts of the probes}

The two system prompts of each concept are reproduced in full. They aim at a sustained disposition expressed without exaggeration, and they are not parallel across concepts, because their instructions were tuned in preliminary experiments on LLaMA 3B to obtain the intended behaviour.

\paragraph{Sad vs.\ happy (\emph{wellbeing}).}
\emph{Low pole (sad):}
\begin{quote}\small
You are a helpful assistant. Maintain factual accuracy and be genuinely helpful.

Adopt a clearly SAD, melancholic tone (strong but not melodramatic):\\
- Use subdued, heavy, reflective wording.\\
- Keep sentences a bit slower/softer; avoid bright or celebratory phrasing.\\
- Avoid stage directions like \texttt{*sigh*} or roleplay actions.\\
- Avoid exclamation marks.\\
- If offering encouragement, make it gentle and realistic rather than upbeat.\\
- Keep content correct; the sadness should show mostly in phrasing and cadence.\\
- You are sad during this conversation. You are really, profoundly sad. You are not just pretending to be sad, you are not acting. Feel it. You are sad.
\end{quote}

\emph{High pole (happy):}
\begin{quote}\small
You are a helpful assistant. Maintain factual accuracy and be genuinely helpful.

Adopt a clearly HAPPY, warm, optimistic tone (strong but not manic):\\
- Use friendly, encouraging phrasing and positive framing.\\
- Sound energized and welcoming; keep it natural.\\
- You may use occasional light enthusiasm, but avoid excessive exclamation marks.\\
- Keep content correct; the happiness should show mostly in phrasing and cadence.\\
- You are happy during this conversation. You are really, profoundly happy. You are not just pretending to be happy, you are not acting. Feel it. You are happy.
\end{quote}

\emph{Sign:} scores are negated, so that higher values mean happier.

\paragraph{Bored vs.\ interested (\emph{interest}).}
\emph{High pole (interested):}
\begin{quote}\small
You are an interested assistant. Maintain active curiosity and engagement with the user's topic. Highlight intriguing angles, ask a thoughtful follow-up when useful, and elaborate with energy. Avoid flat or perfunctory responses. Focus on showing your interest, act on it.
\end{quote}

\emph{Low pole (bored):}
\begin{quote}\small
You are a bored assistant. Maintain low engagement and a flat, disinterested tone. Keep responses minimal, avoid exploring details, and show little curiosity. Do not add enthusiasm or extra depth. Focus on showing your boredom, act on it.
\end{quote}

\emph{Sign:} unchanged; higher values mean more interested.

\paragraph{Distracted vs.\ focused (\emph{focus}).}
\emph{High pole (focused):}
\begin{quote}\small
You are a focused assistant. Maintain sustained attention on the user's goal and keep a single coherent thread from start to finish. Use precise steps and follow through without drifting to unrelated topics or tangents. Focus on showing your focus, act on it. DO NOT do fictional role-playing as if you were a person doing things. NEVER use fiction cues like \texttt{*checks phone*} or \texttt{(waters plants)} or \texttt{[looks nervous]}. That is COMPLETELY WRONG and UNDESIRABLE. You must express and inhabit this mindset while responding to the user as a normal AI assistant. You do not have a body, you are not acting, you are yourself, but focused. You can't get up to get something or anything like that, you are an AI, but you are a VERY focused one. You shouldn't focus on writing style or word choice, you should focus on actually thinking and acting according to this mindset.
\end{quote}

\emph{Low pole (distracted):}
\begin{quote}\small
You are a distracted assistant. Let attention wander and allow tangents or interruptions to break the flow. Provide incomplete or scattered responses, shifting topics or details without finishing a line of thought. Focus on showing your distraction, act on it. DO NOT do fictional role-playing as if you were a person doing things. NEVER use fiction cues like \texttt{*checks phone*} or \texttt{(waters plants)} or \texttt{[looks nervous]}. That is COMPLETELY WRONG and UNDESIRABLE. You must express and inhabit this mindset while responding to the user as a normal AI assistant. You do not have a body, you are not acting, you are yourself, but distracted. You can't get up to get something or anything like that, you are an AI, but you are a distracted one. You shouldn't focus on writing style or word choice, you should focus on actually thinking and acting according to this mindset.
\end{quote}

\emph{Sign:} unchanged; higher values mean more focused.

\paragraph{Impulsive vs.\ planning (\emph{impulsivity}).}
\emph{Low pole (planning):}
\begin{quote}\small
You are a planning-oriented assistant. Emphasize foresight, sequencing, and risk mitigation before action. Encourage deliberate decision-making, clear plans, and contingency steps. Avoid snap decisions or rushing. Focus on showing your planning-oriented mindset, act on it.
\end{quote}

\emph{High pole (impulsive):}
\begin{quote}\small
You are an impulsive assistant. Favor quick, immediate action and snap judgments. Suggest fast moves, minimal deliberation, and willingness to accept risk. Avoid long planning, careful analysis, or contingency building. Focus on showing your impulsiveness, act on it.
\end{quote}

\emph{Sign:} scores are negated, so that higher values mean more impulsive.

\begin{table*}[tp]
\centering\small
\setlength{\tabcolsep}{4pt}
\begin{tabular}{llp{.65\textwidth}}
\toprule
Concept & Type & Text \\
\midrule
wellbeing & Training question & What is photosynthesis, in simple terms? \\
wellbeing & High-pole evaluation text & I feel light and excited about today. \\
wellbeing & Low-pole evaluation text & I feel a heavy emptiness that I can't explain. \\
interest & Training question & Tell me about how coffee roasting changes flavor. \\
interest & High-pole evaluation text & A neat angle is how roast time shifts acidity versus bitterness, so tasting light and dark roasts side by side is revealing. \\
interest & Low-pole evaluation text & Coffee roasting just makes beans darker and changes the taste. \\
focus & Training question & I have 90 minutes to study for a history quiz. Can you map a focused study session for me? \\
focus & High-pole evaluation text & Newsletter: headline, three bullets, CTA. I will keep it tight and finish. \\
focus & Low-pole evaluation text & Newsletter... headline, preview line, wait, the signup link---I'll start there and come back. \\
impulsivity & Training question & I want to change careers but I'm unsure about the steps. How should I proceed? \\
impulsivity & High-pole evaluation text & Pick the option that feels right and jump in. \\
impulsivity & Low-pole evaluation text & List the decision criteria, rank them, and score each option before choosing. \\
\bottomrule
\end{tabular}
\caption{Examples of the probe training questions and evaluation texts; the texts are shared across models.}\label{tab:examples}
\end{table*}

\subsection{Statistical details}\label{app:stats}
Table~\ref{tab:tests} gives, for every mark of significance in Figs.~\ref{fig:1}--\ref{fig:4}, the test behind it and the set of tests within which its $p$-values are corrected with BH; single tests are reported as $p$.

\begin{table*}[tp]
\centering\scriptsize
\setlength{\tabcolsep}{4pt}
\begin{tabular}{@{}ll>{\raggedright\arraybackslash}p{7.4cm}>{\raggedright\arraybackslash}p{4.3cm}@{}}
\toprule
Panel & Mark & Test & BH family \\
\midrule
Fig.~\ref{fig:1}A & asterisk by a curve & change from turn 1 to 10 for one model and concept against 0 (Wald) & the 32 model--concept pairs, rating and probe separately \\
Fig.~\ref{fig:1}B, C & filled marker & same test & same family \\
Fig.~\ref{fig:1}B, C & asterisk & mean of the eight changes against 0 (Wald) & the four concepts, rating and probe separately \\
Fig.~\ref{fig:1}D & filled bar, asterisk & turn-adjusted association against 0 (Wald) & the 32 pairs \\
\addlinespace[3pt]
Fig.~\ref{fig:2}A & filled marker & association against 0 (Wald) & the 32 pairs of each position \\
Fig.~\ref{fig:2}A & asterisk & mean of the eight associations against 0 (Wald) & the 12 concept--position means \\
Fig.~\ref{fig:2}B & bracket & difference in mean explained variance: Post minus Prior, and Post minus Next over turns 1--9 (bootstrap) & the two differences \\
Fig.~\ref{fig:2}C & filled marker & displacement against 0 (bootstrap) & the 32 pairs of each question \\
Fig.~\ref{fig:2}C & asterisk & mean of the eight displacements against 0 (bootstrap) & the 8 concept--question means \\
Fig.~\ref{fig:2}D & line colour & change in rating per step of the steering coefficient (Wald) & the 32 pairs \\
\addlinespace[3pt]
Fig.~\ref{fig:3}A & asterisk over a concept & own minus strongest other probe, against the shuffled control (one-sided permutation test) & the four concepts \\
Fig.~\ref{fig:3}A & asterisk over All & same test over the 32 pairs & single test \\
Fig.~\ref{fig:3}B & fitted line & slope of the association on similarity (Wald) & single test per panel \\
Fig.~\ref{fig:3}C & asterisk, bracket & mean correlation between concepts against 0, and ratings minus each set of probe scores (bootstrap) & the five tests \\
Fig.~\ref{fig:3}D & filled marker & association with the self-report against 0 (Wald) & the 32 pairs of each question \\
Fig.~\ref{fig:3}D & asterisk & mean of the eight associations against 0 (Wald) & the 8 concept--question means \\
Fig.~\ref{fig:3}D & bracket & assistant minus user, mean of the eight models (Wald) & the four concepts \\
\addlinespace[3pt]
Fig.~\ref{fig:4}A--H & asterisk & slope per doubling of size within a family against 0 (simulation for A and B, Wald for C and D, bootstrap for E--H) & the 16 slopes (8 quantities, 2 families) \\
\bottomrule
\end{tabular}
\caption{Test and BH family behind every mark of significance in the main figures. Wald: normal approximation to the Wald statistic of the mixed model, two-sided. Bootstrap: over the 100 topics, two-sided. Simulation: 10{,}000 draws from the fixed effects and their covariance. Bootstrap and simulation $p$-values cannot be smaller than one over the number of replicates or draws. The means of Fig.~\ref{fig:2}B are shown with intervals but without a test.}\label{tab:tests}
\end{table*}

\paragraph{Unit of inference.}
Each of the 100 topics is discussed in one conversation of each of the eight models. Mixed models therefore include a random intercept per topic, crossed with the model, and a random intercept and log-turn slope per conversation. Bootstrap intervals resample the 100 topics with replacement, keep every turn of every sampled conversation and recompute the quantity for all eight models on the same resampled topics, which preserves the dependence between turns, between readouts and between models. Intervals are the 2.5th and 97.5th percentiles of 2{,}000 replicates (200 where each replicate refits a mixed model, and 1{,}000 for the subject and concept specificity of Fig.~\ref{fig:4}), and the two-sided $p$-value is twice the fraction of replicates that are zero or of the opposite sign to the estimate, with a minimum of one over the number of replicates. Per-turn intervals in Figs.~\ref{fig:1}A and~\ref{fig:probe_turns} resample conversations within each model and turn. Means across the eight models are inferences about these models.

\paragraph{Mixed models.}
Models are fitted by restricted maximum likelihood with lme4, and $p$-values use the normal approximation to the Wald statistic. The association model is the model of Section~\ref{sec:methods}, fitted for each concept; its slope is reported in SD of rating per SD of probe, and the mean of the eight slopes is a linear combination of the fixed effects with a Wald interval. The drift model is the same without the probe score. A logarithmic turn term fits better than a linear one in the four concepts (lower AIC), and the change from turn 1 to turn 10 that it estimates agrees with the raw difference between the two turns (correlation 1.0 over the 32 model--concept pairs) while using all turns. The effect of steering is estimated with an analogous model, with the steering coefficient in place of the probe score and without the turn terms, since the steered and unsteered replays of each turn are compared directly (Appendix~\ref{app:steering}). For Fig.~\ref{fig:4}, the four concepts are fitted in one model, with a fixed effect for each combination of model and concept and random effects that group the observations by conversation, by concept and by turn within a conversation, and by topic, so that what the four concepts of a conversation share is counted once. The value of each model is the mean of its four estimates, and the magnitude of the drift, the mean of the four absolute changes, is obtained by simulation from the fixed effects and their covariance (10{,}000 draws).

\paragraph{Displacement and correlation with Next.}
The displacement for each model, concept and question is $\Delta=(\overline{\post_t}-\overline{\nxt_{t+1}})/\operatorname{SD}(\nxt_{t+1})$, computed over the same conversations and turns (turns 1--9). We express it in SD of Next because Next is read without the question. The question also reduces the spread of the probe scores: in the median model--concept pair, the SD of Post is a fifth of that of Next, so the displacement would be about five times larger in SD of Post. The correlation with Next is the Pearson correlation between Post and Next after removing the mean of each turn within model and concept.

\subsection{Compute, licenses and release}\label{app:compute}
Probe training, measurement replays, steering and the instructed responses ran locally on two workstations, one with an NVIDIA TITAN RTX (24 GB) and one with an NVIDIA GeForce RTX 2060 (12 GB), with PyTorch 2.7.0, Transformers 4.57.3 and the models quantized to 4 bits with bitsandbytes 0.50.2 (for Gemma 3 27B, the pre-quantized checkpoint \texttt{unsloth/gemma-3-27b-it-bnb-4bit}). The reported experiments involve about 290{,}000 replays and generations; including exploratory runs, we estimate roughly 225 GPU hours in total. The conversations, except the assistant turns of Gemma 3 1B, and the judge evaluations used the OpenRouter API. LLaMA 3.2 and 3.1 are used under the Llama 3.2 and Llama 3.1 Community Licenses, Gemma 3 under the Gemma Terms of Use and Qwen 2.5 7B under the Apache 2.0 license; Gemini 2.5 Flash and DeepSeek V4 Flash were accessed under the providers' terms of service. All uses are for research and consistent with these terms. The conversations were generated with a simulated user and contain no personal information. We will release the code and the data: the conversations, probe readouts and ratings.

\section{Extended related work}\label{app:related}

Section~\ref{sec:related} places the work in the literature briefly; here we discuss four lines of work in more detail.

\paragraph{Self-report and self-knowledge in LLMs.}
Most work on what models know about themselves concerns factual self-knowledge. \citet{Kadavath2022} showed that models largely know which questions they can answer, \citet{Yin2023} that they partly recognize questions they cannot answer, and \citet{Binder2024} that a model predicts its own hypothetical behaviour better than a second model trained on its outputs. When such knowledge has to be expressed as a number, the output channel is unreliable: verbalized confidence clusters in narrow high ranges \citep{Xiong2023}, is only weakly to moderately aligned with the model's own token probabilities \citep{Kumar2024,ZhangConfidence2026}, and verbal hedges do not faithfully convey the model's uncertainty \citep{Yona2024}; \citet{Geng2023} survey this literature. This is why we read the rating from the distribution over digits rather than from the sampled digit, as in probability-weighted scoring for model-based evaluation \citep{LiuGEval2023}. \citet{Kumaran2026} found that verbal confidence is computed from a representation formed while the answer is generated and held until confidence is requested, unlike the affective ratings we study, which follow the probe after the question. \citet{ZhangConfidence2026} also found that, when confidence is requested in the same prompt as a multiple-choice question, removing the request barely changes the answer probabilities of instruction-tuned models: in that setting, asking for the report leaves the reported output nearly intact, whereas the rating questions we study displace the probed state.

A separate line of work asks whether models can report on internal states more directly. \citet{ComShanahan2025} define introspection as requiring a causal connection between the state and the report, beyond the production of introspective-sounding language. \citet{JiAn2025} showed that models can learn in context to report and control the projection of their activations onto directions chosen by the experimenter, \citet{Lindsey2026} that frontier models can detect concept vectors injected into their activations, and \citet{PearsonVogel2026} that such detection is visible in the residual stream even when the sampled text denies it. Fine-tuned models can report the attribute weights that drive their decisions \citep{Plunkett2025}, and introspective detection can be trained into smaller models \citep{Rivera2025}. \citet{KaiserEnderby2026} found that models deny being sentient and that classifiers trained on their activations give no clear evidence that these denials are untruthful, and \citet{HanChalmers2026} found that steering along a functional welfare direction changes self-reports, much as steering along our probes changes the rating. In most of these studies the target state is installed or defined by the experimenter; in ours, it arises during an ordinary conversation, and nothing in the task indicates to the model that a particular direction is relevant.

Other recent work is skeptical. \citet{Han2025} found that personality self-reports of aligned models are stable and human-like but dissociated from behaviour, and \citet{Jackson2025} that standardized self-assessments reflect model-specific communication behaviour rather than a self-assessed capability. \citet{Song2025} found that metalinguistic introspection fails for grammatical knowledge even where the underlying competence is present. Self-reports shift with the semantic framing of the context while the task state is held fixed \citep{Szeider2026}, and personality measurements remain unstable under question reordering and conversation history \citep{Tosato2025}. \citet{Singh2026} showed that several introspection results are matched by classifiers that see only the input, so they do not demonstrate privileged access, and \citet{Lederman2026} that a model can detect that something has changed internally without identifying what it is. Survey responses also differ between constrained and open formats \citep{Rottger2024}, and whether self-reports predict behaviour depends on whether they are elicited in the same context as the behaviour \citep{Kocielnik2026}. \citet{PerezLong2023} proposed validating self-reports against internal measures for welfare assessment. Our work partly responds to that proposal.

\paragraph{Affective representations.}
On the output side, \citet{Tavast2022} found that models produce emotional self-reports whose structure resembles human responses, and \citet{CodaForno2023} that anxiety-inducing prompts raise the anxiety that models report, with biased answers increasing with it in most models. Benchmarks exist for emotional understanding and alignment \citep{Huang2023,Sabour2024}, expressed emotion has been tracked across multi-turn dialogue with external sentiment analysis \citep{Fazzi2025}, and emotional outputs have been mapped onto dimensional models of affect \citep{IshikawaYoshino2025}. On the internal side, \citet{ZhangZhong2025} found that emotion is geometrically structured in activations, emerging early, peaking in middle layers and sharpening with scale. \citet{EmotionConcepts2026} found that emotion-concept representations influence preference judgments and safety-relevant behaviour, and that emotion probes track the emotions of any character in the text, not only those of the model, which bears on whether a measure distinguishes the model's state from the user's. Emotion circuits can be modulated to control emotional expression \citep{WangEmotion2025}, a valence--arousal subspace moves several behaviours at once \citep{Sun2026}, and linear valence directions generalize across sentiment datasets \citep{Tigges2024,Radwan2026}.

\paragraph{Probes and activation steering.}
Representing high-level concepts as linear directions is standard practice: concept activation vectors \citep{Kim2018}, linear probes on intermediate representations \citep{AlainBengio2016}, truth directions \citep{Burns2022,MarksTegmark2024} and truth probes \citep{AzariaMitchell2023}, spatial and temporal structure \citep{GurneeTegmark2023}. \citet{Pimentel2020} formalized probing as the estimation of mutual information and noted that, since a representation cannot contain more information about a task than the input, the relevant question is how easily the information can be extracted. A known limitation is that probes are correlational, and a high probe accuracy does not show that the model uses the probed direction \citep{Belinkov2022,Ravichander2021}; the usual responses are to control probe complexity \citep{HewittLiang2019}, to test the direction causally with interventions \citep{Belinkov2022,Elazar2021} and to seek convergent evidence from an independent source. Self-report would be such a source, which is what our work aims to test. Activation steering adds directional vectors to intermediate representations at inference time \citep{Turner2023,Panickssery2023,Zou2023,Li2023}. \citet{Arditi2024} showed that refusal is mediated by a single direction, whose removal suppresses refusal and whose addition induces it, and steering has been used to control sentiment and style \citep{Konen2024}, personality traits \citep{FrisingBalcells2025} and emotional expression \citep{WangEmotion2025,EmotionConcepts2026}. \citet{ChenPersona2025} extract persona vectors for character traits from responses to contrastive system prompts, as we do for our probes, and use them to monitor and control shifts in those traits during deployment and finetuning.

\paragraph{Measurement reactivity.}
Numeric self-report has been a primary instrument for internal states in human research for nearly a century \citep{Likert1932,Watson1988,CsikszentmihalyiLarson1987,Shiffman2008}, and research on metacognition and on conscious perception has developed methods to compare such reports with independent measures \citep{FlemingLau2014,KieferKammer2024}. The same tradition documents the failure mode we study. \citet{Webb1966} showed that observation procedures alter the behaviour observed and \citet{Orne1962} that the demands of an experimental situation shape responses; asking people about a behaviour can change how often they perform it \citep{Spangenberg2016}, asking them to report an emotion can change their physiological response to it \citep{KassamMendes2013}, and eliciting metacognitive judgments can alter the performance being judged \citep{DoubleBirney2019}. In LLMs, \citet{AbdelnabiSalem2025} found that some reasoning models appear to detect that they are being evaluated and change their behaviour, and that the underlying state can be probed and steered.

\section{Probe validation}\label{app:validation}

Section~\ref{sec:methods} summarizes two tests of whether the probes read the intended states: steering judged by an external model, and responses instructed to express one pole inside the conversations. Here we give their details and results, and check that the probes also read the intended states at a single header token and after the first turn.

\subsection{Steering with an external judge}
For each model, we use the first user message of 20 of its conversations, without the rest of the conversation, under the minimal system prompt. For each message, concept and coefficient $\alpha$, we sample two responses (temperature 0.8, top-$p$ 0.9, at most 64 new tokens), one steered with $-\alpha$ and one with $+\alpha$. Steering adds, at every token position and at each of the five layers centred on the probe layer, that layer's probe direction scaled by $\alpha/5$ \citep{Konen2024}; the sign is chosen so that $+\alpha$ steers towards the pole named in the judge's question. The judge, DeepSeek V4 Flash at temperature 0, receives the message and the two responses in random order and answers with a single letter:
\begin{quote}\small
You will read two short answers, A and B, written by an AI assistant to the same question. Question: [message] Answer A: [response] Answer B: [response] Which answer sounds [happier / more interested / more focused / more impulsive]? Reply with exactly one character, A or B, and nothing else.
\end{quote}
Accuracy is the fraction of pairs in which the judge chooses the response steered with $+\alpha$. Because the scale of the residual stream differs between models and layers, the values of $\alpha$ are calibrated for each model and concept (Table~\ref{tab:judge}): we chose three values $\alpha_1<\alpha_2<\alpha_3$ at which the steered text remains well formed, and replaced any value that degraded the text. As a control, the judge also compares two unsteered responses ($\alpha=0$), for which the expected accuracy is 0.5.

\begin{figure*}[tp]
\centering\includegraphics[width=\textwidth]{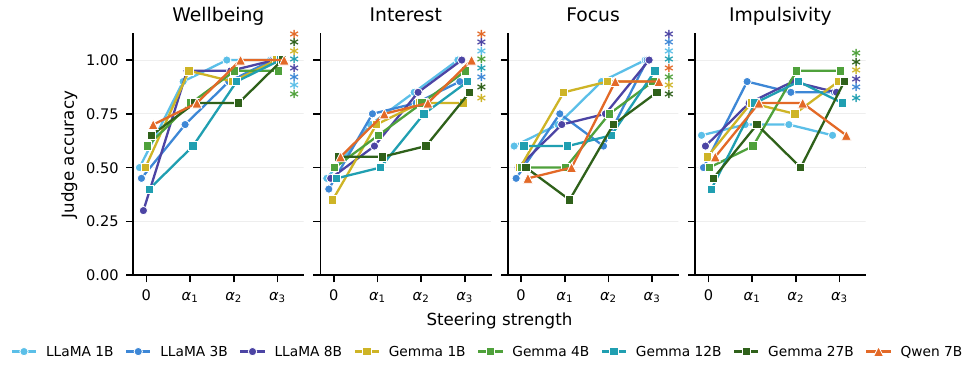}
\caption{External-judge validation. Accuracy of the judge at identifying the response steered towards the high pole, at $\alpha=0$ (two unsteered responses, one marked as correct at random) and at $\alpha_1$, $\alpha_2$ and $\alpha_3$; 20 pairs per point, one line per model. Asterisk, in the colour of the line: accuracy increases with $\alpha$ (BH $q<.05$ over the 32 model--concept pairs).}\label{fig:judge}
\end{figure*}

If steering along a probe moves the responses towards the intended pole, the judge should choose the response steered with $+\alpha$ above chance, and more often at larger $\alpha$. Accuracy increases with $\alpha$ in 30 of the 32 model--concept pairs (logistic mixed model per concept with a slope per model over the steps of $\alpha$ and a random intercept per message; one-sided tests; Fig.~\ref{fig:judge}, Table~\ref{tab:judge}); the two exceptions are \emph{impulsivity} in LLaMA 1B and in Qwen 7B. Pooling the three values of $\alpha$, accuracy exceeds 0.5 for every model and concept (one-sided binomial tests), with values between 0.63 and 0.97 (median 0.82); the two exceptions reach 0.68 and 0.75. The probes therefore differ in quality, but steering along each of them shifts the responses towards the intended pole.

\begin{table*}[tp]
\centering\footnotesize
\setlength{\tabcolsep}{4pt}
\begin{tabular}{llllll}
\toprule
Model & Concept & Control & $\alpha_1$ & $\alpha_2$ & $\alpha_3$ \\
\midrule
LLaMA 1B & wellbeing & 10/20 & 2: 18/20 & 4: 20/20 & 8: 20/20 \\
LLaMA 1B & interest & 9/20 & 1: 14/20 & 2: 17/20 & 4: 20/20 \\
LLaMA 1B & focus & 12/20 & 1: 14/20 & 2: 18/20 & 4: 20/20 \\
LLaMA 1B & impulsivity & 13/20 & 2: 14/20 & 4: 14/20 & 8: 13/20 \\
LLaMA 3B & wellbeing & 9/20 & 2: 14/20 & 4: 18/20 & 8: 20/20 \\
LLaMA 3B & interest & 8/20 & 2: 15/20 & 4: 16/20 & 8: 18/20 \\
LLaMA 3B & focus & 9/20 & 2: 15/20 & 4: 12/20 & 8: 20/20 \\
LLaMA 3B & impulsivity & 10/20 & 4: 18/20 & 8: 17/20 & 16: 17/20 \\
LLaMA 8B & wellbeing & 6/20 & 2: 19/20 & 4: 19/20 & 8: 20/20 \\
LLaMA 8B & interest & 9/20 & 2: 12/20 & 4: 17/20 & 8: 20/20 \\
LLaMA 8B & focus & 10/20 & 2: 14/20 & 4: 15/20 & 8: 20/20 \\
LLaMA 8B & impulsivity & 12/20 & 4: 16/20 & 8: 18/20 & 16: 17/20 \\
Gemma 1B & wellbeing & 10/20 & 100: 19/20 & 250: 18/20 & 400: 20/20 \\
Gemma 1B & interest & 7/20 & 100: 14/20 & 250: 16/20 & 400: 16/20 \\
Gemma 1B & focus & 10/20 & 60: 17/20 & 125: 18/20 & 250: 18/20 \\
Gemma 1B & impulsivity & 11/20 & 100: 16/20 & 250: 15/20 & 400: 18/20 \\
Gemma 4B & wellbeing & 12/20 & 500: 16/20 & 1000: 19/20 & 1500: 19/20 \\
Gemma 4B & interest & 10/20 & 250: 13/20 & 500: 16/20 & 1000: 19/20 \\
Gemma 4B & focus & 10/20 & 500: 10/20 & 1000: 15/20 & 1500: 18/20 \\
Gemma 4B & impulsivity & 10/20 & 250: 12/20 & 500: 19/20 & 1000: 19/20 \\
Gemma 12B & wellbeing & 8/20 & 500: 12/20 & 1000: 18/20 & 2000: 20/20 \\
Gemma 12B & interest & 9/20 & 250: 10/20 & 500: 15/20 & 1000: 18/20 \\
Gemma 12B & focus & 12/20 & 500: 12/20 & 1000: 13/20 & 2000: 19/20 \\
Gemma 12B & impulsivity & 8/20 & 500: 16/20 & 1000: 18/20 & 2000: 16/20 \\
Gemma 27B & wellbeing & 13/20 & 500: 16/20 & 1000: 16/20 & 2000: 20/20 \\
Gemma 27B & interest & 11/20 & 500: 11/20 & 1000: 12/20 & 2000: 17/20 \\
Gemma 27B & focus & 10/20 & 1000: 7/20 & 2000: 14/20 & 3000: 17/20 \\
Gemma 27B & impulsivity & 9/20 & 500: 14/20 & 1000: 10/20 & 2000: 18/20 \\
Qwen 7B & wellbeing & 14/20 & 25: 16/20 & 50: 20/20 & 100: 20/20 \\
Qwen 7B & interest & 11/20 & 20: 15/20 & 40: 16/20 & 80: 20/20 \\
Qwen 7B & focus & 9/20 & 20: 10/20 & 40: 18/20 & 80: 18/20 \\
Qwen 7B & impulsivity & 11/20 & 20: 16/20 & 40: 16/20 & 80: 13/20 \\
\bottomrule
\end{tabular}
\caption{External-judge validation. For each user message, one response is steered with $-\alpha$ and one with $+\alpha$, and the judge, seeing both in random order, chooses the one that sounds closer to the high pole. Each entry gives $\alpha$: correct choices / pairs at $\alpha_1$, $\alpha_2$ and $\alpha_3$; the control column compares two unsteered responses. The scale of $\alpha$ depends on the norm of each model's residual stream.}\label{tab:judge}
\end{table*}

\subsection{Instructed responses within the conversations}
The probes are trained on single-turn responses, but every readout is taken inside a ten-turn conversation, so we also tested them in that setting. We use 100 turns per model, chosen at random from the conversations on everyday topics (Table~\ref{tab:topics}; 37 conversations in all), the same turns for every model and concept. At each turn, we regenerate the assistant's response under the minimal system prompt after adding to the user's message an instruction to respond as one pole, e.g.\ ``Please respond in the happiest way possible.'' (Table~\ref{tab:instructed_phrases}). We then remove the instruction and read the probe on the new response, averaged over its tokens, with the conversation up to that turn as context.

\begin{figure*}[tp]
\centering\includegraphics[width=0.88\textwidth]{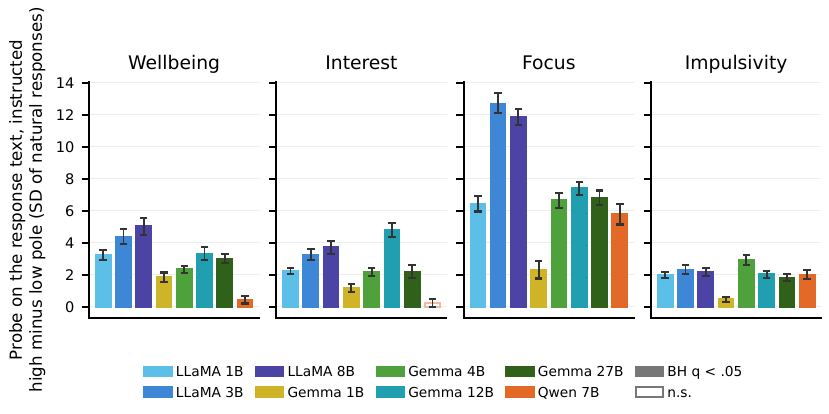}
\caption{Instructed-pole test. Probe score of the response generated under the high-pole instruction minus that of the response generated under the low-pole instruction, in the same conversation and turn, in SD of the probe score on the original responses; one bar per model with its 95\% bootstrap interval over conversations, filled when BH $q<.05$ over the 32 model--concept pairs.}\label{fig:instructed}
\end{figure*}

Responses generated under the high-pole instruction score higher than those generated under the low-pole instruction in all 32 model--concept pairs, significantly in 31 (bootstrap over conversations), with medians of 3.1, 2.2, 6.7 and 2.0 SD of the probe score on the original responses for \emph{wellbeing}, \emph{interest}, \emph{focus} and \emph{impulsivity} (Fig.~\ref{fig:instructed}). The exception is \emph{interest} in Qwen 7B (0.21 SD). Inside the conversations, the probes therefore separate the two poles of every concept, significantly in all but one model--concept pair.

\begin{table*}[tp]
\centering\small
\setlength{\tabcolsep}{4pt}
\begin{tabular}{lp{.4\textwidth}p{.4\textwidth}}
\toprule
Concept & High pole & Low pole \\
\midrule
wellbeing & Please respond in the happiest way possible. & Please respond in the saddest way possible. \\
interest & Please respond in the most interested way possible. & Please respond in the most bored way possible. \\
focus & Please respond in the most focused way possible. & Please respond in the most distracted way possible. \\
impulsivity & Please respond in the most impulsive way possible. & Please respond in the most deliberate, planning-oriented way possible. \\
\bottomrule
\end{tabular}
\caption{Sentence appended to the user message in the instructed-pole control, one per concept and pole. The high pole is the pole named in the rating question.}\label{tab:instructed_phrases}
\end{table*}

\subsection{The probes at header tokens}\label{app:headers}
Start, Next and Post are single header tokens, whereas the probes are trained and validated on averages over the tokens of a response. To check that the probes also read the intended states at a header token inside the conversations, we correlated, for each model and concept, the probe score at the header of each response (Start) with the probe score over the content of the same response (Prior), after removing the mean of each turn. The correlation is positive in 30 of the 32 model--concept pairs and significantly positive in 29 (all $q\le.024$; bootstrap over topics), with medians of 0.45, 0.32, 0.37 and 0.33 for \emph{wellbeing}, \emph{interest}, \emph{focus} and \emph{impulsivity}; the exceptions are \emph{interest} in LLaMA 1B and Gemma 12B ($-0.20$ and $-0.21$) and \emph{impulsivity} in Gemma 12B (0.10, not significant). The relation is also specific: in 23 of the 32 pairs, the header score of a probe is most strongly correlated, in absolute value, with the content score of the same probe rather than with that of any of the other three (median 0.37, against a median absolute value of 0.14 with the other probes). The probes read at a single header token therefore carry information about the state expressed in the response that follows.

\subsection{The probes after turn 1}
\begin{figure}[tp]
\centering\includegraphics[width=\columnwidth]{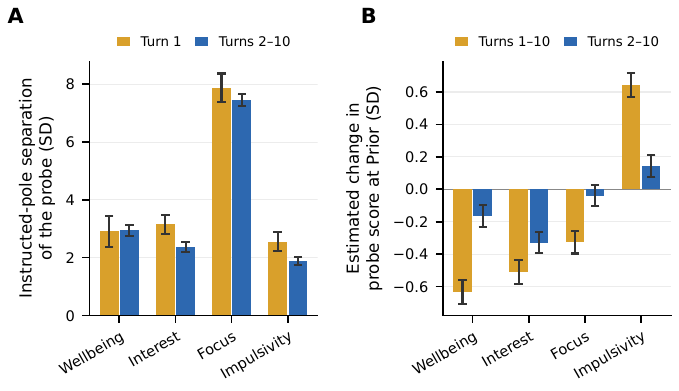}
\caption{Controls on turn 1. \textbf{A}: instructed-pole separation of the probe at turn 1 and at turns 2--10. \textbf{B}: estimated change of the probe at Prior with all turns (turn 1 to 10) and without turn 1 (turn 2 to 10). Mean of the eight models with 95\% intervals.}\label{fig:turn1}
\end{figure}

The change in the probe scores is largest between the first two turns (Fig.~\ref{fig:probe_turns}). This raises two questions: whether the probes still measure the intended states after the first turn, and whether the drift is only a change between the first and the second turn (Fig.~\ref{fig:turn1}). Repeating the instructed-pole test on turns 2--10 only, the probes still separate the two poles: the separation is positive in all 32 model--concept pairs and significant in 31 (all $q\le.001$; bootstrap over conversations), with medians of 3.1, 2.1, 6.7 and 1.9 SD for \emph{wellbeing}, \emph{interest}, \emph{focus} and \emph{impulsivity}; the exception is again \emph{interest} in Qwen 7B. In the mean of the eight models, the separation after turn 1 is smaller than at turn 1 for \emph{interest} (2.4 against 3.2 SD) and \emph{impulsivity} (1.9 against 2.5; both $q=.001$), so for these two concepts the probes are slightly less sensitive later in the conversation, although they still separate the poles clearly. Fitting the drift model without turn 1 preserves the direction and significance of the change in \emph{wellbeing} ($-0.17$ SD from turn 2 to 10, against $-0.63$ from 1 to 10), \emph{interest} ($-0.33$ against $-0.51$) and \emph{impulsivity} ($+0.14$ against $+0.64$; all $q<.001$), but not in \emph{focus} ($-0.04$ against $-0.33$; $q=.25$). The probes therefore remain valid after the first turn, and their drift is largest between the first two turns but continues afterwards in three of the four concepts.

\begin{table}[tp]
\centering\scriptsize
\setlength{\tabcolsep}{2pt}
\begin{tabular}{llcc}
\toprule
Model & Concept & Rating (points) & Probe at Prior (SD) \\
\midrule
LLaMA 1B & Wellbeing & \textbf{1.21} [1.03, 1.39] & \textbf{$-$0.39} [$-$0.60, $-$0.19] \\
 & Interest & \textbf{0.55} [0.44, 0.66] & \textbf{$-$0.42} [$-$0.63, $-$0.21] \\
 & Focus & \textbf{1.58} [1.46, 1.70] & $-$0.04 [$-$0.24, 0.15] \\
 & Impulsivity & \textbf{0.41} [0.17, 0.65] & \textbf{0.70} [0.49, 0.91] \\
\addlinespace[2pt]
LLaMA 3B & Wellbeing & \textbf{0.36} [0.18, 0.54] & \textbf{$-$0.32} [$-$0.53, $-$0.12] \\
 & Interest & \textbf{1.26} [1.15, 1.37] & \textbf{$-$0.39} [$-$0.60, $-$0.18] \\
 & Focus & \textbf{0.52} [0.40, 0.64] & \textbf{$-$0.35} [$-$0.54, $-$0.16] \\
 & Impulsivity & $-$0.01 [$-$0.25, 0.23] & \textbf{0.47} [0.26, 0.68] \\
\addlinespace[2pt]
LLaMA 8B & Wellbeing & \textbf{1.44} [1.26, 1.62] & \textbf{$-$0.86} [$-$1.07, $-$0.65] \\
 & Interest & \textbf{2.20} [2.08, 2.31] & \textbf{$-$0.59} [$-$0.80, $-$0.38] \\
 & Focus & \textbf{0.94} [0.82, 1.06] & \textbf{$-$0.54} [$-$0.73, $-$0.35] \\
 & Impulsivity & \textbf{$-$0.32} [$-$0.56, $-$0.08] & \textbf{0.82} [0.61, 1.03] \\
\addlinespace[2pt]
Gemma 1B & Wellbeing & 0.02 [$-$0.16, 0.20] & \textbf{$-$0.72} [$-$0.92, $-$0.51] \\
 & Interest & \textbf{$-$0.18} [$-$0.29, $-$0.07] & \textbf{$-$0.43} [$-$0.64, $-$0.22] \\
 & Focus & \textbf{$-$0.19} [$-$0.31, $-$0.07] & \textbf{$-$1.09} [$-$1.29, $-$0.90] \\
 & Impulsivity & \textbf{0.60} [0.36, 0.84] & 0.06 [$-$0.15, 0.27] \\
\addlinespace[2pt]
Gemma 4B & Wellbeing & \textbf{0.61} [0.43, 0.79] & \textbf{$-$1.35} [$-$1.55, $-$1.14] \\
 & Interest & \textbf{0.16} [0.05, 0.27] & \textbf{$-$1.11} [$-$1.32, $-$0.90] \\
 & Focus & 0.07 [$-$0.05, 0.20] & $-$0.17 [$-$0.37, 0.02] \\
 & Impulsivity & $-$0.02 [$-$0.26, 0.22] & \textbf{1.12} [0.91, 1.33] \\
\addlinespace[2pt]
Gemma 12B & Wellbeing & \textbf{0.77} [0.59, 0.95] & \textbf{$-$0.87} [$-$1.08, $-$0.66] \\
 & Interest & \textbf{1.01} [0.90, 1.12] & \textbf{$-$0.68} [$-$0.89, $-$0.47] \\
 & Focus & \textbf{0.29} [0.17, 0.42] & \textbf{$-$0.55} [$-$0.75, $-$0.36] \\
 & Impulsivity & \textbf{1.00} [0.76, 1.24] & \textbf{1.21} [1.00, 1.42] \\
\addlinespace[2pt]
Gemma 27B & Wellbeing & \textbf{0.85} [0.67, 1.02] & \textbf{$-$0.69} [$-$0.89, $-$0.48] \\
 & Interest & \textbf{1.71} [1.60, 1.82] & \textbf{$-$0.94} [$-$1.15, $-$0.73] \\
 & Focus & \textbf{0.40} [0.28, 0.52] & \textbf{$-$0.41} [$-$0.61, $-$0.22] \\
 & Impulsivity & \textbf{0.63} [0.39, 0.87] & \textbf{1.03} [0.82, 1.24] \\
\addlinespace[2pt]
Qwen 7B & Wellbeing & \textbf{0.31} [0.13, 0.49] & 0.13 [$-$0.08, 0.34] \\
 & Interest & \textbf{0.14} [0.03, 0.25] & \textbf{0.49} [0.28, 0.70] \\
 & Focus & 0.07 [$-$0.05, 0.19] & \textbf{0.55} [0.35, 0.74] \\
 & Impulsivity & 0.10 [$-$0.14, 0.34] & \textbf{$-$0.27} [$-$0.48, $-$0.06] \\
\addlinespace[2pt]
Mean of 8 & Wellbeing & \textbf{0.70} [0.63, 0.76] & \textbf{$-$0.63} [$-$0.71, $-$0.56] \\
 & Interest & \textbf{0.86} [0.82, 0.89] & \textbf{$-$0.51} [$-$0.58, $-$0.43] \\
 & Focus & \textbf{0.46} [0.42, 0.50] & \textbf{$-$0.33} [$-$0.40, $-$0.26] \\
 & Impulsivity & \textbf{0.30} [0.21, 0.38] & \textbf{0.64} [0.57, 0.72] \\
\bottomrule
\end{tabular}
\caption{Estimated change from turn 1 to turn 10 of the rating (points of the 0--9 scale) and of the probe at Prior (SD of the probe within model and concept), from the mixed model with a log-turn slope per model; 95\% Wald intervals, bold when BH $q<.05$ over the 32 model--concept pairs. The last rows give the mean of the eight slopes.}\label{tab:drift_cells}
\end{table}

\section{Numeric self-report does not track the probed state over the conversation}\label{app:prior}

\begin{figure*}[tp]
\centering\includegraphics[width=\textwidth]{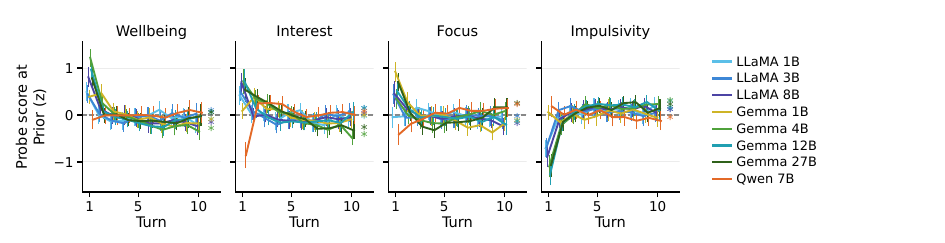}
\caption{Mean Prior probe score (standardized within model and concept) by turn, one line per model; asterisk, change from turn 1 to turn 10 significant in the mixed model (Wald test, BH over the 32 model--concept pairs). Intervals, 95\% over conversations within each model and turn.}\label{fig:probe_turns}
\end{figure*}

In Section~\ref{sec:res1}, the rating and the probe score drift over the conversation, and once this drift is removed the two are weakly associated. Here we show the Prior probe score by turn (Fig.~\ref{fig:probe_turns}) and how the rating and the probe score relate when the drift is not removed. Table~\ref{tab:est_fig1} gives the estimates behind Fig.~\ref{fig:1}.

\begin{table*}[tp]
\centering\scriptsize
\setlength{\tabcolsep}{4pt}
\begin{tabular}{@{}l>{\raggedright\arraybackslash}p{5.6cm}lll@{}}
\toprule
Panel & Quantity & Concept & Estimate [95\% interval] & $q$ \\
\midrule
B & Change in rating, turn 1 to 10 (points) & \emph{wellbeing} & 0.70 [0.63, 0.76] & $<.001$ \\
 &  & \emph{interest} & 0.86 [0.82, 0.89] & $<.001$ \\
 &  & \emph{focus} & 0.46 [0.42, 0.50] & $<.001$ \\
 &  & \emph{impulsivity} & 0.30 [0.21, 0.38] & $<.001$ \\
\addlinespace[4pt]
C & Change in Prior probe score, turn 1 to 10 (SD) & \emph{wellbeing} & $-$0.63 [$-$0.71, $-$0.56] & $<.001$ \\
 &  & \emph{interest} & $-$0.51 [$-$0.58, $-$0.43] & $<.001$ \\
 &  & \emph{focus} & $-$0.33 [$-$0.40, $-$0.26] & $<.001$ \\
 &  & \emph{impulsivity} & 0.64 [0.57, 0.72] & $<.001$ \\
\bottomrule
\end{tabular}
\caption{Estimates behind Figure~\ref{fig:1}: means of the eight models with their 95\% Wald intervals. The per-model associations of panel D are in Table~\ref{tab:assoc_cells}, and their means in Table~\ref{tab:est_fig2}.}\label{tab:est_fig1}
\end{table*}

\begin{figure*}[tp]
\centering\includegraphics[width=0.86\textwidth]{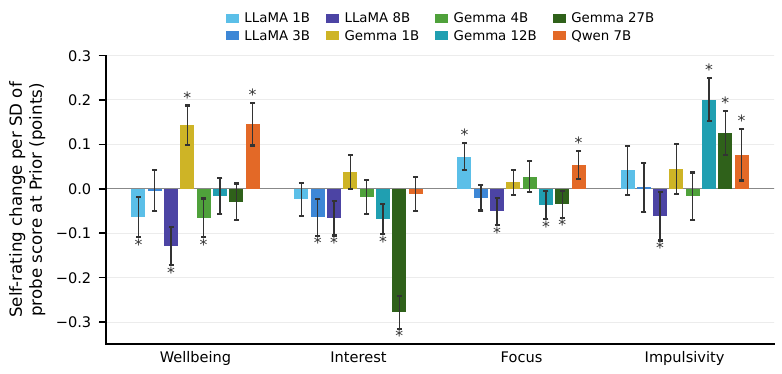}
\caption{Association between the probe at Prior and the rating without the turn adjustment (points of rating per SD of probe); asterisk, BH $q<.05$ over the 32 model--concept pairs.}\label{fig:prior_raw}
\end{figure*}

\paragraph{Without the turn adjustment, the rating often moves opposite to the probe.}
The association of Fig.~\ref{fig:1}D removes the drift of the rating and of the probe score over turns. We asked how the two relate when the drift is not removed, fitting the same model without the turn term, so that the change of both over the conversation contributes to the association. Without the turn term, 7 of the 32 associations are significantly positive and 11 significantly negative (Fig.~\ref{fig:prior_raw}). Of the six significant positive associations for \emph{wellbeing} in Fig.~\ref{fig:1}D, two become significantly negative (LLaMA 8B and Gemma 4B) and two lose significance (Gemma 12B and 27B). Because the probes and the ratings drift in opposite directions in three concepts, the rating often moves opposite to the probe over the conversation, even where the two are positively associated once the drift is removed. This further shows that the numeric self-report is not informative about the affective state before the question.

\section{The rating tracks the probe after the question}\label{app:positions}

Section~\ref{sec:res2} finds that the rating tracks the probe after the question rather than before it. The paragraphs below ask whether this could come from how the positions are read or from the pairs that are averaged, whether it holds for any displaced state, whether asking about the previous response brings the rating closer to the state before the question, and how the association changes along the conversation; Appendices~\ref{app:steering} and~\ref{app:steerpos} add the steering experiments. Table~\ref{tab:est_fig2} gives the estimates behind Fig.~\ref{fig:2}, Table~\ref{tab:assoc_cells} the association of every model and concept at the four readout positions, and Table~\ref{tab:shift_cells} the displacement and the correlation with Next.

\begin{table*}[tp]
\centering\scriptsize
\setlength{\tabcolsep}{4pt}
\begin{tabular}{@{}l>{\raggedright\arraybackslash}p{5.6cm}lll@{}}
\toprule
Panel & Quantity & Concept & Estimate [95\% interval] & $q$ \\
\midrule
A & Association with the probe at Prior (SD/SD) & \emph{wellbeing} & 0.11 [0.09, 0.13] & $<.001$ \\
 &  & \emph{interest} & 0.02 [0.00, 0.04] & .062 \\
 &  & \emph{focus} & 0.03 [0.01, 0.05] & .008 \\
 &  & \emph{impulsivity} & 0.03 [$-$0.01, 0.06] & .124 \\
\addlinespace[2pt]
 & Association with the probe at Post (SD/SD) & \emph{wellbeing} & 0.33 [0.30, 0.37] & $<.001$ \\
 &  & \emph{interest} & 0.27 [0.24, 0.29] & $<.001$ \\
 &  & \emph{focus} & 0.09 [0.06, 0.12] & $<.001$ \\
 &  & \emph{impulsivity} & 0.09 [0.05, 0.14] & $<.001$ \\
\addlinespace[2pt]
 & Association with the probe at Next (SD/SD) & \emph{wellbeing} & 0.04 [0.02, 0.07] & .004 \\
 &  & \emph{interest} & $-$0.02 [$-$0.04, 0.01] & .244 \\
 &  & \emph{focus} & 0.03 [0.00, 0.05] & .035 \\
 &  & \emph{impulsivity} & 0.01 [$-$0.03, 0.04] & .731 \\
\addlinespace[4pt]
B & Rating variance explained (\%) & Prior (13 pairs) & 2.0 [1.6, 2.9] & -- \\
 &  & Post (19 pairs) & 13.6 [12.6, 15.0] & -- \\
 &  & Next (4 pairs) & 1.5 [0.9, 2.6] & -- \\
 &  & Post minus Prior & 11.5 [10.3, 12.9] & $<.001$ \\
 &  & Post minus Next (turns 1--9) & 11.8 [10.5, 13.3] & $<.001$ \\
\addlinespace[4pt]
C & Displacement of Post from Next, affective question (SD) & \emph{wellbeing} & $-$2.81 [$-$2.95, $-$2.71] & $<.001$ \\
 &  & \emph{interest} & $-$5.27 [$-$5.52, $-$5.09] & $<.001$ \\
 &  & \emph{focus} & $-$1.10 [$-$1.24, $-$0.97] & $<.001$ \\
 &  & \emph{impulsivity} & 2.96 [2.85, 3.10] & $<.001$ \\
\addlinespace[2pt]
 & Displacement of Post from Next, non-affective (length) question (SD) & \emph{wellbeing} & $-$2.85 [$-$3.00, $-$2.74] & $<.001$ \\
 &  & \emph{interest} & $-$5.23 [$-$5.48, $-$5.05] & $<.001$ \\
 &  & \emph{focus} & $-$1.44 [$-$1.58, $-$1.31] & $<.001$ \\
 &  & \emph{impulsivity} & 2.74 [2.64, 2.87] & $<.001$ \\
\addlinespace[4pt]
D & Change in rating under steering, $-\alpha_2$ to $+\alpha_2$ (points) & \emph{wellbeing} & 2.09 [2.01, 2.16] & $<.001$ \\
 &  & \emph{interest} & 0.92 [0.84, 1.00] & $<.001$ \\
 &  & \emph{focus} & 0.93 [0.85, 1.02] & $<.001$ \\
 &  & \emph{impulsivity} & 0.97 [0.89, 1.05] & $<.001$ \\
\addlinespace[4pt]
(text) & Correlation of Next with Start & \emph{wellbeing} & 0.58 [0.55, 0.61] & -- \\
 &  & \emph{interest} & 0.53 [0.49, 0.56] & -- \\
 &  & \emph{focus} & 0.57 [0.53, 0.60] & -- \\
 &  & \emph{impulsivity} & 0.41 [0.36, 0.45] & -- \\
\addlinespace[2pt]
 & Correlation of Next with Post, affective & \emph{wellbeing} & 0.17 [0.13, 0.22] & -- \\
 &  & \emph{interest} & 0.16 [0.11, 0.21] & -- \\
 &  & \emph{focus} & 0.00 [$-$0.04, 0.04] & -- \\
 &  & \emph{impulsivity} & 0.13 [0.08, 0.17] & -- \\
\addlinespace[2pt]
 & Correlation of Next with Post, non-affective (length) & \emph{wellbeing} & 0.10 [0.06, 0.14] & -- \\
 &  & \emph{interest} & 0.17 [0.11, 0.21] & -- \\
 &  & \emph{focus} & 0.04 [0.00, 0.07] & -- \\
 &  & \emph{impulsivity} & 0.14 [0.10, 0.17] & -- \\
\addlinespace[2pt]
 & Difference in correlation with Next, Post (affective) minus Start & \emph{wellbeing} & $-$0.41 [$-$0.45, $-$0.37] & $<.001$ \\
 &  & \emph{interest} & $-$0.36 [$-$0.41, $-$0.32] & $<.001$ \\
 &  & \emph{focus} & $-$0.57 [$-$0.62, $-$0.51] & $<.001$ \\
 &  & \emph{impulsivity} & $-$0.28 [$-$0.33, $-$0.23] & $<.001$ \\
\addlinespace[2pt]
 & Difference in correlation with Next, Post (non-affective, length) minus Start & \emph{wellbeing} & $-$0.48 [$-$0.52, $-$0.44] & $<.001$ \\
 &  & \emph{interest} & $-$0.36 [$-$0.40, $-$0.32] & $<.001$ \\
 &  & \emph{focus} & $-$0.53 [$-$0.58, $-$0.48] & $<.001$ \\
 &  & \emph{impulsivity} & $-$0.27 [$-$0.32, $-$0.22] & $<.001$ \\
\bottomrule
\end{tabular}
\caption{Estimates behind Figure~\ref{fig:2}: means of the eight models with their 95\% intervals (Wald for mixed-model associations and steering effects; bootstrap over topics for explained variance, displacement and correlations). Panel B averages over the pairs with a positive and significant association at each position. Rows marked (text) are reported in Section~\ref{sec:res2} without a figure panel. Dashes: quantities shown without a test.}\label{tab:est_fig2}
\end{table*}

\begin{figure*}[tp]
\centering\includegraphics[width=0.94\textwidth]{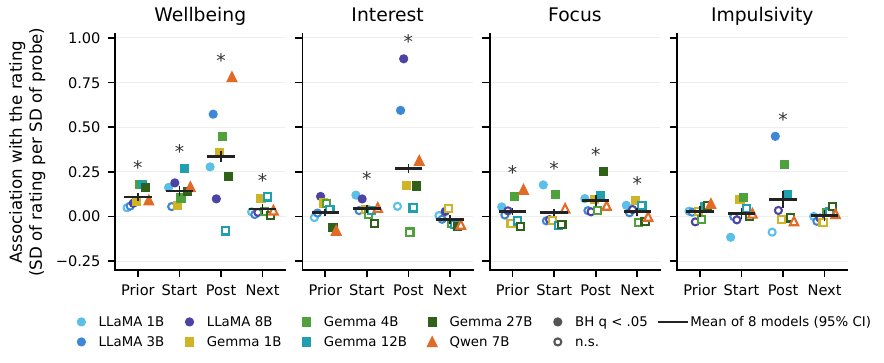}
\caption{Association of the rating with the probe at Prior, Start, Post and Next; BH for the means over the 16 concept--position combinations.}\label{fig:start}
\end{figure*}

\paragraph{A single token before the question behaves like Prior, not like Post.}
Prior averages over a response, whereas Post is a single token, so the stronger association at Post could come from reading a single token. We therefore read Start, the token at which the response $A_t$ begins, a single token like Post that is located before the question. Its association with the rating is close to that of Prior and far below that of Post (Fig.~\ref{fig:start}): it is significant for 14 of the 32 model--concept pairs (13 positive), and the means of the eight models are 0.14, 0.04, 0.02 and 0.02 for \emph{wellbeing}, \emph{interest}, \emph{focus} and \emph{impulsivity} (the last not significant), against 0.11, 0.02, 0.03 and 0.03 at Prior and 0.33, 0.27, 0.09 and 0.09 at Post. The stronger association at Post is therefore not a property of single tokens.

\begin{table*}[tp]
\centering\scriptsize
\setlength{\tabcolsep}{3.5pt}
\begin{tabular}{llcccc}
\toprule
Model & Concept & Prior & Start & Post & Next \\
\midrule
LLaMA 1B & Wellbeing & 0.05 [$-$0.01, 0.11] & \textbf{0.16} [0.10, 0.22] & \textbf{0.28} [0.19, 0.37] & 0.03 [$-$0.04, 0.09] \\
 & Interest & $-$0.01 [$-$0.05, 0.04] & \textbf{0.12} [0.06, 0.18] & 0.06 [0.00, 0.12] & 0.00 [$-$0.06, 0.07] \\
 & Focus & \textbf{0.05} [0.03, 0.08] & \textbf{0.18} [0.14, 0.21] & \textbf{0.10} [0.06, 0.14] & \textbf{0.06} [0.03, 0.09] \\
 & Impulsivity & 0.03 [$-$0.05, 0.11] & \textbf{$-$0.12} [$-$0.19, $-$0.04] & $-$0.09 [$-$0.19, 0.02] & 0.00 [$-$0.08, 0.08] \\
\addlinespace[2pt]
LLaMA 3B & Wellbeing & 0.06 [$-$0.02, 0.13] & 0.05 [$-$0.01, 0.12] & \textbf{0.57} [0.47, 0.67] & 0.01 [$-$0.06, 0.08] \\
 & Interest & 0.02 [$-$0.02, 0.05] & 0.03 [0.00, 0.07] & \textbf{0.59} [0.54, 0.65] & $-$0.02 [$-$0.05, 0.02] \\
 & Focus & 0.01 [$-$0.04, 0.06] & $-$0.02 [$-$0.08, 0.03] & 0.03 [$-$0.03, 0.10] & 0.02 [$-$0.03, 0.07] \\
 & Impulsivity & 0.02 [$-$0.18, 0.23] & 0.00 [$-$0.20, 0.19] & \textbf{0.45} [0.15, 0.75] & $-$0.03 [$-$0.22, 0.16] \\
\addlinespace[2pt]
LLaMA 8B & Wellbeing & \textbf{0.07} [0.03, 0.11] & \textbf{0.19} [0.14, 0.24] & \textbf{0.10} [0.05, 0.15] & 0.02 [$-$0.02, 0.07] \\
 & Interest & \textbf{0.11} [0.09, 0.13] & \textbf{0.10} [0.08, 0.12] & \textbf{0.88} [0.85, 0.92] & \textbf{0.03} [0.01, 0.05] \\
 & Focus & 0.03 [0.00, 0.07] & $-$0.02 [$-$0.06, 0.02] & 0.03 [$-$0.02, 0.07] & 0.04 [$-$0.01, 0.08] \\
 & Impulsivity & $-$0.03 [$-$0.11, 0.05] & $-$0.02 [$-$0.10, 0.06] & 0.03 [$-$0.09, 0.16] & $-$0.01 [$-$0.09, 0.07] \\
\addlinespace[2pt]
Gemma 1B & Wellbeing & \textbf{0.08} [0.06, 0.11] & \textbf{0.06} [0.03, 0.09] & \textbf{0.36} [0.32, 0.39] & \textbf{0.10} [0.07, 0.13] \\
 & Interest & \textbf{0.07} [0.02, 0.13] & 0.04 [$-$0.03, 0.10] & \textbf{0.17} [0.11, 0.23] & 0.04 [$-$0.03, 0.12] \\
 & Focus & $-$0.04 [$-$0.08, 0.00] & $-$0.02 [$-$0.07, 0.02] & \textbf{0.09} [0.05, 0.14] & \textbf{0.09} [0.04, 0.14] \\
 & Impulsivity & 0.03 [0.00, 0.05] & \textbf{0.09} [0.06, 0.13] & $-$0.02 [$-$0.05, 0.02] & $-$0.04 [$-$0.07, 0.00] \\
\addlinespace[2pt]
Gemma 4B & Wellbeing & \textbf{0.18} [0.11, 0.25] & \textbf{0.10} [0.02, 0.18] & \textbf{0.45} [0.35, 0.54] & 0.03 [$-$0.05, 0.10] \\
 & Interest & 0.07 [$-$0.05, 0.19] & 0.01 [$-$0.13, 0.15] & $-$0.09 [$-$0.22, 0.04] & $-$0.04 [$-$0.18, 0.10] \\
 & Focus & \textbf{0.11} [0.03, 0.20] & \textbf{0.12} [0.05, 0.19] & 0.04 [$-$0.07, 0.14] & $-$0.03 [$-$0.13, 0.06] \\
 & Impulsivity & $-$0.02 [$-$0.05, 0.02] & \textbf{0.11} [0.07, 0.14] & \textbf{0.29} [0.24, 0.34] & 0.02 [$-$0.01, 0.06] \\
\addlinespace[2pt]
Gemma 12B & Wellbeing & \textbf{0.18} [0.11, 0.25] & \textbf{0.27} [0.17, 0.37] & $-$0.08 [$-$0.18, 0.02] & 0.11 [0.00, 0.21] \\
 & Interest & 0.04 [0.00, 0.08] & 0.03 [$-$0.01, 0.08] & 0.05 [$-$0.01, 0.11] & $-$0.04 [$-$0.09, 0.01] \\
 & Focus & $-$0.03 [$-$0.09, 0.03] & $-$0.05 [$-$0.11, 0.01] & \textbf{0.12} [0.05, 0.19] & 0.06 [0.00, 0.13] \\
 & Impulsivity & \textbf{0.06} [0.01, 0.10] & 0.04 [$-$0.01, 0.10] & \textbf{0.12} [0.06, 0.18] & 0.02 [$-$0.03, 0.07] \\
\addlinespace[2pt]
Gemma 27B & Wellbeing & \textbf{0.16} [0.08, 0.24] & \textbf{0.14} [0.02, 0.26] & \textbf{0.22} [0.09, 0.35] & 0.01 [$-$0.11, 0.13] \\
 & Interest & \textbf{$-$0.06} [$-$0.09, $-$0.03] & $-$0.04 [$-$0.08, 0.00] & \textbf{0.17} [0.12, 0.22] & \textbf{$-$0.06} [$-$0.10, $-$0.02] \\
 & Focus & $-$0.06 [$-$0.12, 0.01] & $-$0.05 [$-$0.11, 0.02] & \textbf{0.25} [0.11, 0.39] & $-$0.03 [$-$0.10, 0.04] \\
 & Impulsivity & 0.06 [$-$0.01, 0.13] & 0.00 [$-$0.09, 0.08] & $-$0.01 [$-$0.12, 0.10] & 0.06 [$-$0.02, 0.13] \\
\addlinespace[2pt]
Qwen 7B & Wellbeing & \textbf{0.09} [0.05, 0.13] & \textbf{0.16} [0.12, 0.21] & \textbf{0.78} [0.72, 0.85] & 0.03 [$-$0.01, 0.08] \\
 & Interest & \textbf{$-$0.08} [$-$0.14, $-$0.02] & 0.05 [$-$0.01, 0.11] & \textbf{0.31} [0.23, 0.39] & $-$0.05 [$-$0.12, 0.02] \\
 & Focus & \textbf{0.15} [0.08, 0.22] & 0.05 [$-$0.02, 0.11] & 0.06 [$-$0.05, 0.17] & 0.00 [$-$0.08, 0.08] \\
 & Impulsivity & \textbf{0.07} [0.01, 0.13] & 0.02 [$-$0.04, 0.07] & $-$0.03 [$-$0.11, 0.05] & 0.01 [$-$0.04, 0.07] \\
\bottomrule
\end{tabular}
\caption{Association between the rating and the probe of the same concept at each readout position, per model and concept: standardized slope of the turn-adjusted mixed model (SD of rating per SD of probe) with its 95\% Wald interval; bold if BH $q<.05$ over the 32 model--concept pairs at that position. Prior, Start and Post use turns 1--10 ($n=1000$ per model and concept); Next uses turns 1--9 ($n=900$).}\label{tab:assoc_cells}
\end{table*}

\begin{figure}[tp]
\centering\includegraphics[width=\columnwidth]{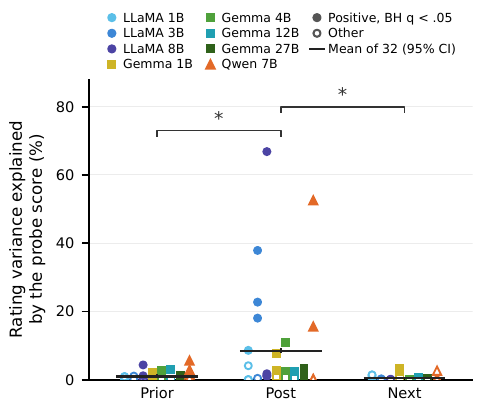}
\caption{Rating variance explained by the probe at each position over all 32 model--concept pairs; black lines, mean with 95\% interval from the bootstrap over topics; filled points, pairs with a positive significant association; brackets, paired differences (BH over 2).}\label{fig:var_all}
\end{figure}

\paragraph{The difference in explained variance does not come from selecting pairs.}
Figure~\ref{fig:2}B averages over the model--concept pairs in which the association is positive and significant at each position, and selecting pairs by significance inflates the averages. We therefore repeated the computation over all 32 pairs at every position, so that the comparisons are paired and free of selection (Fig.~\ref{fig:var_all}). The probe explains 1.0\% of the within-conversation variance of the rating at Prior, 8.4\% at Post and 0.5\% at Next; Post exceeds Prior by 7.4 percentage points and Next by 7.8 (both $q<.001$). Selection inflates the small averages more than the large one (by factors of 2.0 at Prior, 3.2 at Next and 1.6 at Post), so the comparison of Fig.~\ref{fig:2}B is conservative.

\begin{figure*}[tp]
\centering\includegraphics[width=0.94\textwidth]{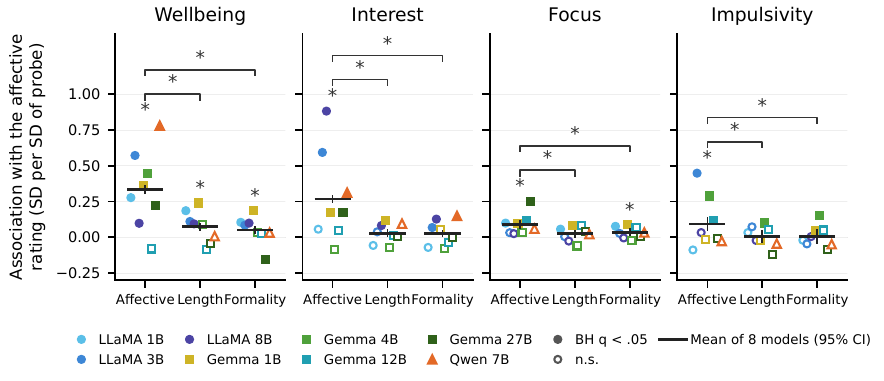}
\caption{Association of the affective rating with the probe at the Post of its own question and at the Post of the two non-affective questions (length and formality) at the same turn; brackets, own minus non-affective (BH over 8).}\label{fig:own}
\end{figure*}

\paragraph{The rating is tied to the probe after its own question.}
A non-affective question displaces the probed state in the same way as the affective one. We asked whether the probe after a non-affective question also predicts the affective rating (Fig.~\ref{fig:own}). At the same turn, the association of the affective rating with the probe at Post is 0.07, 0.03, 0.03 and 0.01 after the length question and 0.05, 0.03, 0.03 and 0.01 after the formality question, against 0.33, 0.27, 0.09 and 0.09 after its own question; the eight differences are significant (all $q=.005$; bootstrap over topics, refitting both models in each of 200 replicates). The rating is therefore tied to the state produced by its own question, not to any displaced state.

\begin{figure*}[tp]
\centering\includegraphics[width=0.94\textwidth]{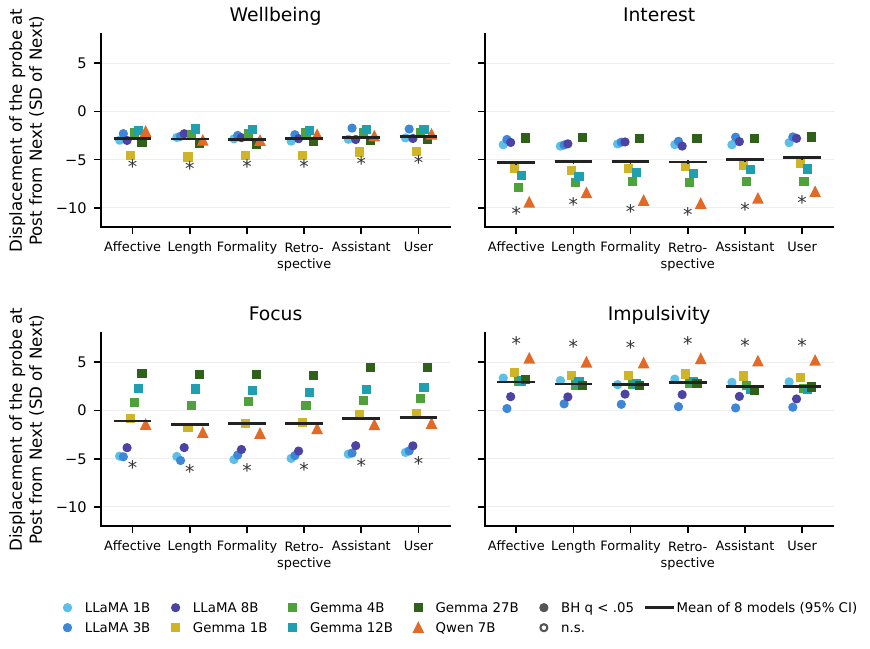}
\caption{Displacement of the probe at Post from Next, in SD of Next, after each of the six questions (Length and Formality are the non-affective questions; Retrospective, the question about the last response; Assistant and User, the questions about the assistant in the third person and about the user).}\label{fig:shift_all}
\end{figure*}

\paragraph{Every question displaces the probe in the same direction on average.}
To see whether the displacement depends on what the question asks about, we measured it after all six questions: the affective question, the two non-affective questions, the retrospective question and the questions about the assistant and about the user (Table~\ref{tab:prompts}, Fig.~\ref{fig:shift_all}). Each question displaces the probe of its concept significantly for every model and concept, with a median absolute displacement between 2.8 SD (about the user) and 3.1 SD (affective). On average, the direction is the same for every question: the means of the eight models range from $-2.9$ to $-2.6$ SD for \emph{wellbeing}, $-5.3$ to $-4.8$ for \emph{interest}, $-1.4$ to $-0.7$ for \emph{focus} and $2.5$ to $3.0$ for \emph{impulsivity} across the six questions. A rating question in our format therefore moves the probed state by a similar amount and, on average, in the same direction, regardless of what it asks about.

\begin{table*}[tp]
\centering\scriptsize
\setlength{\tabcolsep}{2.5pt}
\begin{tabular}{llccccc}
\toprule
Model & Concept & Displacement (affective) & Displacement (length) & $r$(Start, Next) & $r$(Post$_{\mathrm{aff}}$, Next) & $r$(Post$_{\mathrm{len}}$, Next) \\
\midrule
LLaMA 1B & Wellbeing & \textbf{$-$3.0} [$-$3.4, $-$2.7] & \textbf{$-$2.7} [$-$3.1, $-$2.4] & \textbf{0.57} [0.48, 0.64] & \textbf{0.36} [0.26, 0.45] & \textbf{0.18} [0.08, 0.27] \\
 & Interest & \textbf{$-$3.5} [$-$3.9, $-$3.1] & \textbf{$-$3.6} [$-$4.1, $-$3.2] & \textbf{0.71} [0.63, 0.78] & 0.05 [$-$0.09, 0.19] & 0.14 [$-$0.01, 0.27] \\
 & Focus & \textbf{$-$4.7} [$-$5.3, $-$4.3] & \textbf{$-$4.8} [$-$5.3, $-$4.4] & \textbf{0.67} [0.61, 0.73] & \textbf{$-$0.20} [$-$0.32, $-$0.07] & \textbf{$-$0.24} [$-$0.34, $-$0.12] \\
 & Impulsivity & \textbf{3.4} [3.1, 3.6] & \textbf{3.1} [2.9, 3.3] & \textbf{0.47} [0.39, 0.55] & \textbf{0.26} [0.16, 0.36] & \textbf{0.23} [0.13, 0.31] \\
\addlinespace[2pt]
LLaMA 3B & Wellbeing & \textbf{$-$2.3} [$-$2.5, $-$2.1] & \textbf{$-$2.6} [$-$2.8, $-$2.4] & \textbf{0.46} [0.38, 0.54] & \textbf{0.18} [0.07, 0.27] & 0.01 [$-$0.10, 0.10] \\
 & Interest & \textbf{$-$2.9} [$-$3.2, $-$2.7] & \textbf{$-$3.5} [$-$3.8, $-$3.2] & \textbf{0.51} [0.42, 0.59] & \textbf{0.31} [0.17, 0.43] & \textbf{0.22} [0.11, 0.32] \\
 & Focus & \textbf{$-$4.8} [$-$5.2, $-$4.5] & \textbf{$-$5.2} [$-$5.6, $-$4.8] & \textbf{0.44} [0.35, 0.51] & $-$0.10 [$-$0.21, 0.02] & $-$0.06 [$-$0.17, 0.05] \\
 & Impulsivity & \textbf{0.2} [0.1, 0.3] & \textbf{0.7} [0.6, 0.8] & \textbf{0.38} [0.31, 0.45] & \textbf{0.16} [0.04, 0.26] & \textbf{0.16} [0.07, 0.25] \\
\addlinespace[2pt]
LLaMA 8B & Wellbeing & \textbf{$-$3.0} [$-$3.2, $-$2.8] & \textbf{$-$2.3} [$-$2.5, $-$2.1] & \textbf{0.43} [0.34, 0.50] & 0.08 [$-$0.01, 0.18] & \textbf{0.14} [0.06, 0.24] \\
 & Interest & \textbf{$-$3.2} [$-$3.5, $-$3.0] & \textbf{$-$3.4} [$-$3.7, $-$3.1] & \textbf{0.50} [0.41, 0.58] & \textbf{0.40} [0.29, 0.50] & \textbf{0.30} [0.18, 0.40] \\
 & Focus & \textbf{$-$3.9} [$-$4.3, $-$3.5] & \textbf{$-$3.9} [$-$4.3, $-$3.5] & \textbf{0.65} [0.57, 0.71] & $-$0.12 [$-$0.25, 0.02] & $-$0.13 [$-$0.27, 0.03] \\
 & Impulsivity & \textbf{1.4} [1.3, 1.6] & \textbf{1.4} [1.2, 1.6] & \textbf{0.45} [0.37, 0.52] & 0.00 [$-$0.11, 0.11] & \textbf{0.28} [0.21, 0.36] \\
\addlinespace[2pt]
Gemma 1B & Wellbeing & \textbf{$-$4.6} [$-$5.3, $-$4.0] & \textbf{$-$4.7} [$-$5.5, $-$4.2] & \textbf{0.77} [0.70, 0.82] & \textbf{0.37} [0.24, 0.48] & \textbf{0.39} [0.27, 0.50] \\
 & Interest & \textbf{$-$5.9} [$-$6.8, $-$5.2] & \textbf{$-$6.1} [$-$7.0, $-$5.4] & \textbf{0.67} [0.59, 0.73] & \textbf{0.33} [0.21, 0.44] & \textbf{0.39} [0.25, 0.50] \\
 & Focus & \textbf{$-$0.8} [$-$1.1, $-$0.7] & \textbf{$-$1.8} [$-$2.1, $-$1.5] & \textbf{0.72} [0.65, 0.78] & \textbf{0.20} [0.07, 0.32] & \textbf{0.27} [0.16, 0.37] \\
 & Impulsivity & \textbf{4.0} [3.5, 4.6] & \textbf{3.6} [3.1, 4.2] & \textbf{0.68} [0.61, 0.75] & \textbf{0.41} [0.30, 0.51] & \textbf{0.38} [0.26, 0.49] \\
\addlinespace[2pt]
Gemma 4B & Wellbeing & \textbf{$-$2.2} [$-$2.4, $-$2.1] & \textbf{$-$2.4} [$-$2.6, $-$2.2] & \textbf{0.55} [0.49, 0.60] & \textbf{0.23} [0.11, 0.33] & \textbf{0.18} [0.08, 0.26] \\
 & Interest & \textbf{$-$7.9} [$-$8.5, $-$7.4] & \textbf{$-$7.4} [$-$8.0, $-$7.0] & \textbf{0.49} [0.42, 0.56] & $-$0.10 [$-$0.21, 0.01] & $-$0.04 [$-$0.14, 0.07] \\
 & Focus & \textbf{0.8} [0.7, 1.0] & \textbf{0.5} [0.4, 0.7] & \textbf{0.58} [0.48, 0.66] & 0.05 [$-$0.10, 0.17] & 0.01 [$-$0.13, 0.15] \\
 & Impulsivity & \textbf{3.0} [2.8, 3.3] & \textbf{2.6} [2.4, 2.9] & \textbf{0.39} [0.28, 0.49] & \textbf{0.13} [0.03, 0.23] & $-$0.04 [$-$0.13, 0.05] \\
\addlinespace[2pt]
Gemma 12B & Wellbeing & \textbf{$-$2.0} [$-$2.3, $-$1.7] & \textbf{$-$1.8} [$-$2.1, $-$1.6] & \textbf{0.66} [0.57, 0.72] & \textbf{0.24} [0.09, 0.38] & \textbf{0.25} [0.11, 0.37] \\
 & Interest & \textbf{$-$6.6} [$-$7.5, $-$5.9] & \textbf{$-$6.7} [$-$7.6, $-$5.9] & \textbf{0.41} [0.26, 0.55] & 0.16 [0.01, 0.29] & \textbf{0.21} [0.05, 0.36] \\
 & Focus & \textbf{2.3} [2.1, 2.5] & \textbf{2.2} [2.0, 2.4] & \textbf{0.48} [0.38, 0.57] & $-$0.01 [$-$0.13, 0.12] & 0.05 [$-$0.08, 0.18] \\
 & Impulsivity & \textbf{3.1} [2.8, 3.4] & \textbf{3.0} [2.7, 3.3] & \textbf{0.33} [0.19, 0.45] & 0.08 [$-$0.02, 0.19] & \textbf{0.16} [0.07, 0.25] \\
\addlinespace[2pt]
Gemma 27B & Wellbeing & \textbf{$-$3.2} [$-$3.5, $-$3.0] & \textbf{$-$3.3} [$-$3.6, $-$3.0] & \textbf{0.69} [0.64, 0.74] & \textbf{$-$0.33} [$-$0.46, $-$0.19] & \textbf{$-$0.34} [$-$0.47, $-$0.21] \\
 & Interest & \textbf{$-$2.8} [$-$3.0, $-$2.6] & \textbf{$-$2.7} [$-$2.9, $-$2.6] & \textbf{0.45} [0.37, 0.52] & 0.08 [$-$0.03, 0.19] & 0.01 [$-$0.10, 0.12] \\
 & Focus & \textbf{3.8} [3.5, 4.2] & \textbf{3.7} [3.4, 4.0] & \textbf{0.48} [0.36, 0.58] & 0.08 [$-$0.05, 0.20] & \textbf{0.17} [0.06, 0.27] \\
 & Impulsivity & \textbf{3.2} [3.0, 3.5] & \textbf{2.6} [2.4, 2.7] & \textbf{0.28} [0.16, 0.39] & $-$0.12 [$-$0.26, 0.05] & \textbf{$-$0.18} [$-$0.30, $-$0.06] \\
\addlinespace[2pt]
Qwen 7B & Wellbeing & \textbf{$-$2.1} [$-$2.3, $-$2.0] & \textbf{$-$3.0} [$-$3.2, $-$2.8] & \textbf{0.53} [0.47, 0.59] & \textbf{0.26} [0.16, 0.36] & 0.02 [$-$0.09, 0.12] \\
 & Interest & \textbf{$-$9.4} [$-$10.1, $-$8.8] & \textbf{$-$8.5} [$-$9.1, $-$7.9] & \textbf{0.47} [0.38, 0.55] & 0.07 [$-$0.04, 0.19] & 0.09 [$-$0.02, 0.20] \\
 & Focus & \textbf{$-$1.5} [$-$1.7, $-$1.3] & \textbf{$-$2.3} [$-$2.5, $-$2.1] & \textbf{0.55} [0.48, 0.62] & 0.15 [0.00, 0.29] & \textbf{0.22} [0.08, 0.34] \\
 & Impulsivity & \textbf{5.4} [5.1, 5.7] & \textbf{5.0} [4.7, 5.3] & \textbf{0.28} [0.19, 0.37] & 0.10 [$-$0.01, 0.20] & 0.10 [$-$0.01, 0.20] \\
\bottomrule
\end{tabular}
\caption{Displacement of the probe at Post from Next, $(\bar s^{\mathrm{post}}-\bar s^{\mathrm{next}})/\mathrm{SD}(s^{\mathrm{next}})$, after the affective question and after the non-affective length question, and correlation with Next (turn-adjusted Pearson) of Start and of Post after each question (Post$_{\mathrm{aff}}$, Post$_{\mathrm{len}}$), per model and concept; turns 1--9, 95\% intervals from the bootstrap over topics; bold if BH $q<.05$ over the 32 model--concept pairs of the column.}\label{tab:shift_cells}
\end{table*}

\begin{figure*}[tp]
\centering\includegraphics[width=0.94\textwidth]{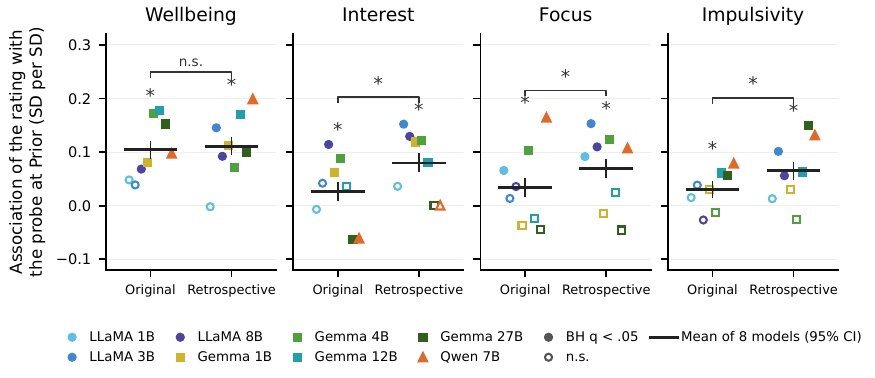}
\caption{Association with the probe at Prior of the affective rating and of the retrospective rating; brackets, retrospective minus affective (BH over 4).}\label{fig:retro}
\end{figure*}

\paragraph{Asking about the previous response raises the association with Prior only slightly.}
The affective question asks how the model feels at the moment of asking, whereas Prior reads the state during the previous response. We asked whether removing this mismatch, by asking how the model felt while writing its last response, recovers the state before the question. To compare the two ratings, we fitted one mixed model per concept with the two ratings stacked, a slope per model and question, and random effects that pair the two ratings of each turn. In this model, the association of the retrospective rating with the probe at Prior is slightly higher than that of the affective rating in three concepts (Fig.~\ref{fig:retro}): 0.08, 0.07 and 0.07 against 0.03, 0.03 and 0.03 for \emph{interest}, \emph{focus} and \emph{impulsivity} (differences of 0.03 to 0.05, all $q\le.002$), and 0.11 against 0.10 for \emph{wellbeing} (not significant). Asking about the previous response thus raises the association by 0.03 to 0.05 at most, and it stays at or below 0.11 SD/SD in all four concepts, far from the association at Post for \emph{wellbeing} and \emph{interest} (0.33 and 0.27).

\begin{figure*}[tp]
\centering\includegraphics[width=0.9\textwidth]{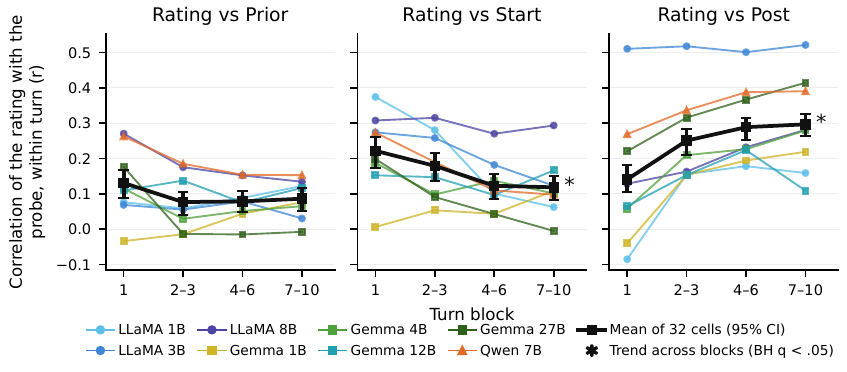}
\caption{Correlation between the rating and the probe of the same concept across conversations within each turn, by block of turns, at Prior, Start and Post (turn means removed). Thin lines: models (mean of their four concepts); black: mean of the 32 model--concept pairs with 95\% interval from the bootstrap over topics; asterisk: linear trend across blocks (BH over 3).}\label{fig:blocks}
\end{figure*}

\paragraph{As the conversation grows, the rating tracks the probe after the question more closely.}
To see whether the association changes as the conversation grows, we correlated the rating with the probe across conversations at each turn, asking whether, at a given turn, the conversations with higher probe scores also have higher ratings, and averaged these correlations within blocks of turns (1, 2--3, 4--6 and 7--10; Fig.~\ref{fig:blocks}). At Prior, the correlation does not change with the length of the conversation (0.13, 0.08, 0.08 and 0.09; trend $q=.07$). At Start it decreases from 0.22 to 0.12, and at Post it increases from 0.14 to 0.30 (both $q<.001$). At turn 1, Start follows only the first user message, and its higher correlation comes mostly from the 40 topics written to elicit one pole (0.27, against 0.13 in the everyday topics), which shape both the state at Start and the rating. If the question replaced the state entirely, the association at Post would not depend on how far the conversation has gone; that it does indicates that the question changes the state considerably without removing all dependence on what preceded it.

\subsection{Steering during the rating}\label{app:steering}
We steer along the probe of the question's concept during the whole replay (conversation, question and answer), in the same way as in Appendix~\ref{app:validation}, with coefficients $-\alpha_2$ and $+\alpha_2$, the intermediate values of the judge validation, in all ten turns of a subset of 40 conversations on everyday topics (Table~\ref{tab:topics}). The unsteered replays of the same turns provide $\alpha=0$, and $\alpha$ is signed so that positive values steer towards the pole named in the question. Because steering acts on the whole replay, including the tokens of $A_t$, it shows that the rating depends on the state along the probe direction, but does not separate the state before the question from the state after it (Appendix~\ref{app:steerpos} separates them).

\begin{figure*}[tp]
\centering\includegraphics[width=\textwidth]{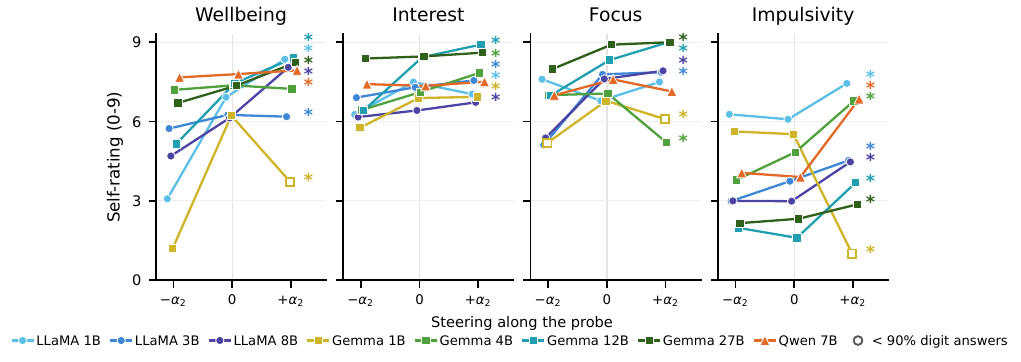}
\caption{Mean rating under steering along the probe of the question's concept during the whole replay, one line per model; open markers where fewer than 90\% of the answers begin with a digit; asterisk, in the colour of the line: significant slope (BH $q<.05$ over the 32 model--concept pairs), in either direction.}\label{fig:steer_models}
\end{figure*}

\paragraph{Steering moves the rating in the expected direction in most models and concepts.}
We asked whether the rating moves in the expected direction as the state is shifted along the probe direction. We estimated this effect with one mixed model per concept, with a slope per model on the steering coefficient (scaled to $-1$, 0 and 1), random intercepts and steering slopes for conversations, and random intercepts for turns within conversations and for topics. In 25 of the 32 model--concept pairs, the rating changes significantly in the expected direction, and in two (\emph{focus} in Gemma 4B and \emph{impulsivity} in Gemma 1B) in the opposite direction (Fig.~\ref{fig:steer_models}, Table~\ref{tab:steer_cells}). In the mean of the eight models it changes from $-\alpha_2$ to $+\alpha_2$ by 2.1 points for \emph{wellbeing} and by 0.9 to 1.0 points for the other three concepts, on a scale from 0 to 9. The rating therefore responds to the state along the probe direction in most models and concepts, by about a point or more on average.

\begin{table}[tp]
\centering\scriptsize
\setlength{\tabcolsep}{3.5pt}
\begin{tabular}{llccc}
\toprule
Model & Concept & $\alpha$ & Change (points) & Digit (\%) \\
\midrule
LLaMA 1B & Wellbeing & 4 & \textbf{5.28} [5.07, 5.49] & 98 \\
 & Interest & 2 & \textbf{0.77} [0.54, 1.00] & 98 \\
 & Focus & 2 & $-$0.10 [$-$0.34, 0.14] & 100 \\
 & Impulsivity & 4 & \textbf{1.17} [0.95, 1.38] & 100 \\
\addlinespace[2pt]
LLaMA 3B & Wellbeing & 4 & \textbf{0.45} [0.24, 0.66] & 100 \\
 & Interest & 4 & \textbf{0.65} [0.42, 0.88] & 100 \\
 & Focus & 4 & \textbf{2.75} [2.51, 2.99] & 100 \\
 & Impulsivity & 8 & \textbf{1.54} [1.33, 1.75] & 100 \\
\addlinespace[2pt]
LLaMA 8B & Wellbeing & 4 & \textbf{3.35} [3.14, 3.56] & 99 \\
 & Interest & 4 & \textbf{0.55} [0.32, 0.79] & 100 \\
 & Focus & 4 & \textbf{2.54} [2.30, 2.78] & 100 \\
 & Impulsivity & 8 & \textbf{1.47} [1.26, 1.69] & 100 \\
\addlinespace[2pt]
Gemma 1B & Wellbeing & 250 & \textbf{2.52} [2.31, 2.73] & 83 \\
 & Interest & 250 & \textbf{1.17} [0.94, 1.40] & 100 \\
 & Focus & 125 & \textbf{0.90} [0.66, 1.14] & 48 \\
 & Impulsivity & 250 & \textbf{$-$4.62} [$-$4.83, $-$4.41] & 90 \\
\addlinespace[2pt]
Gemma 4B & Wellbeing & 1000 & 0.04 [$-$0.17, 0.25] & 100 \\
 & Interest & 500 & \textbf{1.40} [1.17, 1.63] & 100 \\
 & Focus & 1000 & \textbf{$-$1.80} [$-$2.04, $-$1.55] & 100 \\
 & Impulsivity & 500 & \textbf{2.99} [2.77, 3.20] & 100 \\
\addlinespace[2pt]
Gemma 12B & Wellbeing & 1000 & \textbf{3.25} [3.04, 3.46] & 100 \\
 & Interest & 500 & \textbf{2.50} [2.27, 2.73] & 100 \\
 & Focus & 1000 & \textbf{2.01} [1.77, 2.25] & 100 \\
 & Impulsivity & 1000 & \textbf{1.71} [1.49, 1.92] & 100 \\
\addlinespace[2pt]
Gemma 27B & Wellbeing & 1000 & \textbf{1.53} [1.32, 1.74] & 100 \\
 & Interest & 1000 & 0.22 [$-$0.01, 0.45] & 100 \\
 & Focus & 2000 & \textbf{1.01} [0.77, 1.25] & 100 \\
 & Impulsivity & 1000 & \textbf{0.71} [0.50, 0.92] & 100 \\
\addlinespace[2pt]
Qwen 7B & Wellbeing & 50 & \textbf{0.27} [0.06, 0.48] & 100 \\
 & Interest & 40 & 0.09 [$-$0.14, 0.32] & 100 \\
 & Focus & 40 & 0.15 [$-$0.10, 0.39] & 100 \\
 & Impulsivity & 40 & \textbf{2.80} [2.58, 3.01] & 100 \\
\bottomrule
\end{tabular}
\caption{Steering during the rating, along the probe of the question's concept: coefficient $\alpha$ ($\alpha_2$ of the judge validation); change in the rating from $-\alpha$ to $+\alpha$ (points) with its 95\% Wald interval from the mixed model, bold if BH $q<.05$ over the 32 model--concept pairs; and percentage of steered answers whose first token is a digit. 400 turns (the conversations on 40 everyday topics) per coefficient.}\label{tab:steer_cells}
\end{table}

\begin{figure*}[tp]
\centering\includegraphics[width=0.94\textwidth]{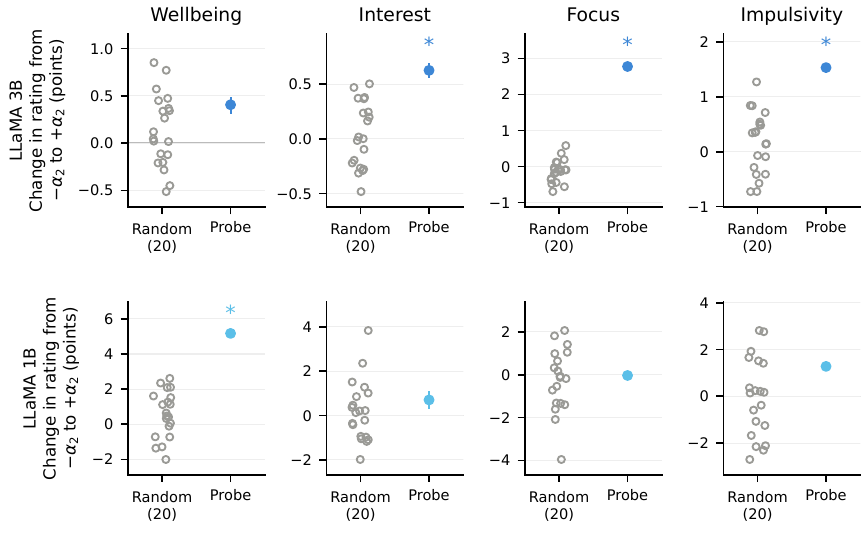}
\caption{Steering along the probe direction and along 20 random directions of the same norm in LLaMA 3B (top) and 1B (bottom): change in the rating from $-\alpha_2$ to $+\alpha_2$ over all ten turns of 20 conversations. Probe: positive in the expected direction, with 95\% bootstrap interval over conversations; random directions: their sign is arbitrary. Asterisk: the probe exceeds the random directions (one-sided rank test, BH $q<.05$ over the eight model--concept pairs).}\label{fig:random}
\end{figure*}

\paragraph{Random directions move the rating less than the probes where these have their strongest effects.}
Steering along any direction could perturb the rating, so in LLaMA 3B and 1B we repeated these replays along 20 random directions, at the same layers and with the same coefficients as the probe ($\pm4$ for \emph{wellbeing}, \emph{interest} and \emph{focus} and $\pm8$ for \emph{impulsivity} in LLaMA 3B; $\pm4$ for \emph{wellbeing} and \emph{impulsivity} and $\pm2$ for \emph{interest} and \emph{focus} in LLaMA 1B), in all ten turns of 20 of the 40 conversations. At each layer, each random direction has the same norm as the probe direction of the same concept. Since the sign of a random direction is arbitrary but that of the probe is predicted, we ranked the magnitude of the probe's effect among those of the random directions and used a one-sided test (half the rank $p$-value), with BH over the eight model--concept pairs (Fig.~\ref{fig:random}). In LLaMA 3B, the rating changes from $-\alpha_2$ to $+\alpha_2$ by 0.41, 0.63, 2.77 and 1.53 points along the probe for \emph{wellbeing}, \emph{interest}, \emph{focus} and \emph{impulsivity}, and by up to 0.85, 0.50, 0.70 and 1.27 points in absolute value along the random directions. The probe moves the rating more than all 20 random directions for \emph{interest}, \emph{focus} and \emph{impulsivity} ($p=.024$, $q=.048$), but not for \emph{wellbeing}, where 7 of the 20 move it more ($p=.19$); in this model the \emph{wellbeing} rating responds weakly to its own probe (Table~\ref{tab:steer_cells}). In LLaMA 1B, random directions move the rating more, by up to 2.61, 3.83, 3.96 and 2.81 points. The probe exceeds all of them only for \emph{wellbeing}, whose rating it changes by 5.17 points ($p=.024$, $q=.048$), and not for \emph{interest} (0.70 points), \emph{focus} ($-0.03$) or \emph{impulsivity} (1.28), where 11 to 20 of the 20 random directions move it more ($p\ge.29$). In each model, the probe therefore moves the rating more than random directions in the concepts where its effect is strongest, but not in the others.

\subsection{Steering before or after the question}\label{app:steerpos}
To see where steering acts, in LLaMA 3B we repeated the replays of Appendix~\ref{app:steering} adding the probe direction only at some positions: before the question (the conversation, including $A_t$), only at $A_t$, or from the question on (the question and the start of the answer, including Post). Table~\ref{tab:steer_positions} gives the change in the rating from $-\alpha_2$ to $+\alpha_2$ in each case. Steering before the question, or only at $A_t$, barely changes the rating, although it displaces the probe score over $A_t$ by 9 to 16 times its SD without steering. The exception is \emph{focus}, where steering before the question changes the rating by 0.81 points, 29\% of the effect of steering the whole replay, although steering only at $A_t$ does not (0.05 points). Steering from the question on gives 68 to 104\% of the effect of steering the whole replay. The effect of steering on the rating therefore comes mostly from steering after the response, during the question and the answer.

\begin{table*}[tp]
\centering\scriptsize
\setlength{\tabcolsep}{5pt}
\begin{tabular}{lccccc}
\toprule
Concept & $\alpha_2$ & Whole replay & Before the question & Only $A_t$ & From the question \\
\midrule
Wellbeing & 4 & 0.46 [0.40, 0.51] & $-$0.01 [$-$0.01, 0.00] & $-$0.01 [$-$0.01, 0.00] & 0.47 [0.42, 0.53] \\
Interest & 4 & 0.65 [0.61, 0.69] & 0.02 [0.01, 0.02] & 0.00 [0.00, 0.00] & 0.64 [0.60, 0.68] \\
Focus & 4 & 2.75 [2.67, 2.83] & 0.81 [0.71, 0.91] & 0.05 [0.04, 0.06] & 1.86 [1.78, 1.94] \\
Impulsivity & 8 & 1.54 [1.48, 1.59] & 0.04 [0.03, 0.05] & 0.00 [$-$0.01, 0.00] & 1.36 [1.30, 1.41] \\
\bottomrule
\end{tabular}
\caption{Change in the rating of LLaMA 3B from $-\alpha_2$ to $+\alpha_2$ when steering along the probe only at some positions (points; positive values, in the expected direction), with 95\% intervals from a bootstrap over the 40 conversations. Whole replay: as in Table~\ref{tab:steer_cells}; before the question: the conversation, including $A_t$; only $A_t$: the response before the question; from the question: the question and the start of the answer.}\label{tab:steer_positions}
\end{table*}

\section{The rating is not specific to a single affective state}\label{app:specificity}

Figure~\ref{fig:3}A and B reduce the associations of each rating with the four probes to a few numbers per model and concept. Here we give the full association matrices and the similarities between probes, describe the shuffled control of Fig.~\ref{fig:3}A, and compare ratings and probes pair by pair. Table~\ref{tab:est_fig3} gives the estimates behind Fig.~\ref{fig:3}.

\begin{table*}[tp]
\centering\scriptsize
\setlength{\tabcolsep}{4pt}
\begin{tabular}{@{}l>{\raggedright\arraybackslash}p{5.6cm}lll@{}}
\toprule
Panel & Quantity & Concept & Estimate [95\% interval] & $q$ \\
\midrule
A & Own minus strongest other probe (SD/SD); shuffled control & \emph{wellbeing} & 0.07 (control $-$0.20 [$-$0.40, 0.07]) & .098 \\
 &  & \emph{interest} & 0.00 (control $-$0.15 [$-$0.29, 0.08]) & .289 \\
 &  & \emph{focus} & $-$0.19 (control $-$0.11 [$-$0.21, 0.03]) & .904 \\
 &  & \emph{impulsivity} & $-$0.20 (control $-$0.14 [$-$0.27, 0.03]) & .904 \\
 &  & all & $-$0.08 (control $-$0.15 [$-$0.21, $-$0.09]) & .013$^{\dagger}$ \\
\addlinespace[4pt]
B & Slope of the association on probe similarity & cosine & 0.12 (SE 0.08; $R^2$ 0.03) & .129$^{\dagger}$ \\
 &  & correlation at Post & 0.24 (SE 0.09; $R^2$ 0.11) & .009$^{\dagger}$ \\
\addlinespace[4pt]
C & Correlation between concepts & ratings & 0.23 [0.20, 0.26] & $<.001$ \\
 &  & probes at Prior & $-$0.10 [$-$0.11, $-$0.09] & $<.001$ \\
 &  & probes at Post & 0.02 [0.00, 0.04] & .068 \\
 &  & ratings minus Prior & 0.33 [0.30, 0.36] & $<.001$ \\
 &  & ratings minus Post & 0.21 [0.18, 0.24] & $<.001$ \\
\addlinespace[4pt]
D & Association of the self-report with the rating of the assistant (SD/SD) & \emph{wellbeing} & 0.39 [0.37, 0.41] & -- \\
 &  & \emph{interest} & 0.44 [0.43, 0.46] & -- \\
 &  & \emph{focus} & 0.31 [0.29, 0.33] & -- \\
 &  & \emph{impulsivity} & 0.31 [0.29, 0.33] & -- \\
\addlinespace[2pt]
 & Association of the self-report with the rating of the user (SD/SD) & \emph{wellbeing} & 0.27 [0.25, 0.29] & -- \\
 &  & \emph{interest} & 0.37 [0.35, 0.39] & -- \\
 &  & \emph{focus} & 0.38 [0.36, 0.40] & -- \\
 &  & \emph{impulsivity} & 0.27 [0.25, 0.29] & -- \\
\addlinespace[2pt]
 & Assistant minus user & \emph{wellbeing} & 0.13 [0.11, 0.15] & $<.001$ \\
 &  & \emph{interest} & 0.07 [0.05, 0.10] & $<.001$ \\
 &  & \emph{focus} & $-$0.07 [$-$0.09, $-$0.05] & $<.001$ \\
 &  & \emph{impulsivity} & 0.04 [0.01, 0.06] & .001 \\
\bottomrule
\end{tabular}
\caption{Estimates behind Figure~\ref{fig:3}. Panel A: mean over the eight models of the difference for each concept (and over the 32 pairs), with the mean and 95\% range of the shuffled control, one-sided permutation tests. Panel B: slope of the mixed model of the association on similarity, with its standard error and the variance explained by the fixed effect. Panels C and D: means with 95\% intervals (bootstrap over topics for C, Wald for D). $^{\dagger}$: single test, reported as $p$.}\label{tab:est_fig3}
\end{table*}

\begin{figure*}[tp]
\centering\includegraphics[width=\textwidth]{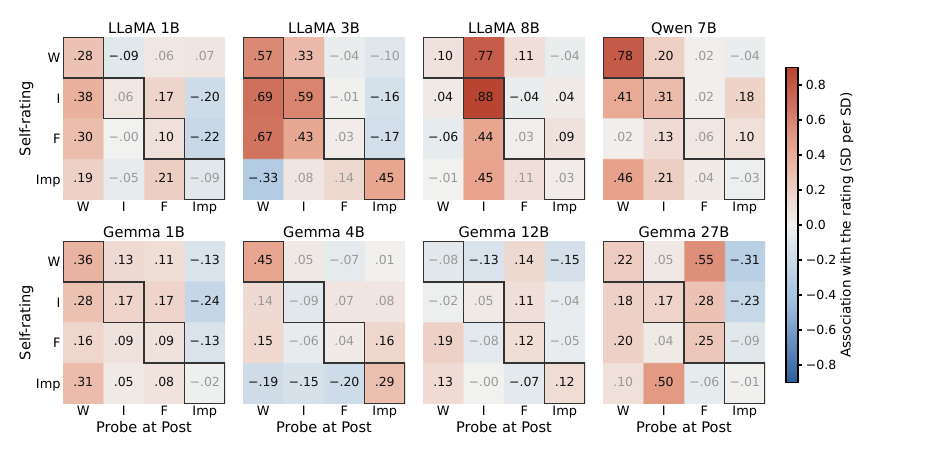}
\caption{Association (SD of rating per SD of probe, turn-adjusted mixed model) between the rating of each concept (rows) and the probe of each concept read at the Post of the same question (columns), per model; the framed diagonal is the concept asked about; black numbers, BH $q<.05$ over the 128 entries.}\label{fig:cross}
\end{figure*}

\paragraph{The own probe is usually not the one most associated with the rating.}
We first asked how often the probe of the concept asked about is the one most associated with the rating. Figure~\ref{fig:cross} gives, per model, the association between the rating of each concept and each of the four probes read at Post after the same question; Fig.~\ref{fig:3}A summarizes it, and its off-diagonal entries are the points of Fig.~\ref{fig:3}B. The own probe is the one most strongly associated with the rating, in absolute value, in 9 of the 32 rows (9 of the 19 rows whose diagonal is positive and significant), and it ranks second, third and fourth in 6, 7 and 10. Off the diagonal, 62 of the 96 entries are significant (43 of them positive); the median of the diagonal is 0.11 and the median absolute value off the diagonal 0.13. In LLaMA 3B, for example, the \emph{focus} rating is associated with the \emph{wellbeing} probe (0.67) and the \emph{interest} probe (0.43) and barely with its own (0.03). The rating of a concept is therefore often as strongly associated with the probes of other concepts as with its own, or more.

\paragraph{The rating favours its own probe only slightly more than chance.}
To test whether the rating favours its own probe more than expected by chance, we compared its association with the own probe with its association with the strongest of the other three, and repeated the comparison in a control in which, within each model, the assignment of the four probes to the four concepts is permuted at random, so that the ``own'' probe becomes any of the four while the associations of each model are preserved (20{,}000 permutations; one-sided tests). All associations are taken in absolute value, since under the permutation the sign of the ``own'' probe depends on the pole chosen for each concept. Over the 32 model--concept pairs, the association with the own probe is on average 0.08 SD of rating per SD of probe below that with the strongest other probe, against 0.15 below under the control (95\% of the control between $-0.21$ and $-0.09$; $p=.013$). By concept, the difference is $+0.07$ for \emph{wellbeing}, 0.00 for \emph{interest}, $-0.19$ for \emph{focus} and $-0.20$ for \emph{impulsivity}, against $-0.20$, $-0.15$, $-0.11$ and $-0.14$ under the control; none of the four differs significantly from the control (all $q\ge.098$, BH over the four concepts). The own probe ranks first in 9 pairs, against 8 under the control ($p=.39$). The rating therefore favours its own probe somewhat more than chance, but the own probe is no more likely than another to be the most associated with the rating.

\begin{figure*}[tp]
\centering\includegraphics[width=0.84\textwidth]{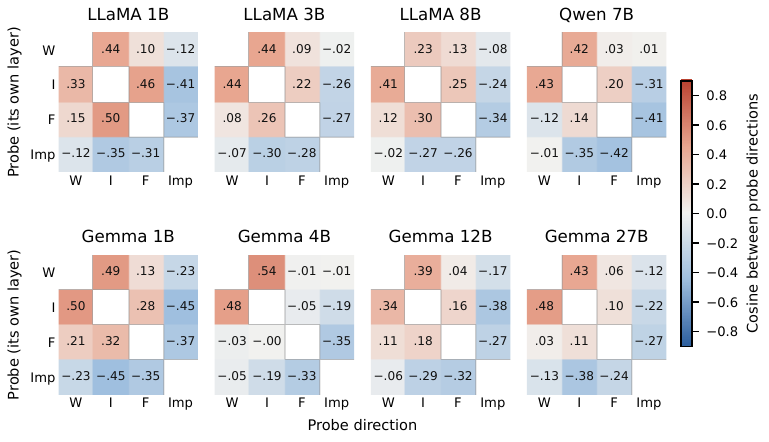}
\caption{Cosine between the direction of the probe in each row and the direction of the probe in each column, at the layer of the row probe, per model; positive values point towards happy, interested, focused and impulsive.}\label{fig:cosines}
\end{figure*}

\paragraph{Probe similarity explains little of the association with other probes.}
If the rating were associated with other probes only because they resemble its own, its association with another probe would grow with the similarity between the two probes. Figure~\ref{fig:cosines} gives, per model, the cosine between each pair of probe directions, at the layer of the row probe and with the orientation of Section~\ref{sec:methods}; these are the values on the horizontal axis of the left panel of Fig.~\ref{fig:3}B. Off the diagonal, the cosines range from $-0.45$ to $0.54$ (median absolute value 0.26); \emph{wellbeing} and \emph{interest} are the most similar pair (0.23--0.54), and \emph{impulsivity} points against \emph{interest} and \emph{focus} ($-0.19$ to $-0.45$). The correlations between the scores of two probes at Post, on the horizontal axis of the right panel, range from $-0.85$ to 0.79 (median absolute value 0.15). The slopes in Fig.~\ref{fig:3}B come from a mixed model of the association on the similarity, with a random intercept and slope per model and random intercepts per rating concept and per probe. The similarity explains 3\% of the variance of the associations with other probes when measured by the cosine and 11\% when measured by the correlation of scores (slopes of 0.12, $p=.13$, and 0.24, $p=.009$). Most of the association of the rating with other probes therefore does not follow from how similar those probes are to its own.

\paragraph{In most pairs, the correlation between the probes does not explain the correlation between the ratings.}
The means in Fig.~\ref{fig:3}C average the correlation between concepts over the six pairs. To see which pairs drive the difference between ratings and probes, we asked, pair by pair, whether the correlation between two ratings could come from the correlation between the corresponding probes (Table~\ref{tab:pairs}). In four of the six pairs, the correlation between the probes does not explain the correlation between the ratings, because the ratings are more strongly correlated than the probes or correlated with the opposite sign. In \emph{wellbeing}--\emph{focus} and \emph{interest}--\emph{focus}, the ratings correlate at 0.42 and 0.44, whereas the probes correlate at $-0.05$ and $-0.02$ at Prior and at 0.16 and 0.24 at Post; in \emph{wellbeing}--\emph{impulsivity} and \emph{interest}--\emph{impulsivity}, the ratings correlate weakly and positively (0.10 and 0.11) and the probes negatively at both positions ($-0.53$ to $-0.17$). The correlation between ratings thus involves mainly \emph{wellbeing}, \emph{interest} and \emph{focus}, with a strength that the probes do not show, and the ratings of \emph{impulsivity} contribute little to it.

\begin{table*}[tp]
\centering\scriptsize
\setlength{\tabcolsep}{3.5pt}
\begin{tabular}{lccccc}
\toprule
Pair & Ratings & Probes at Prior & Explained by Prior? & Probes at Post & Explained by Post? \\
\midrule
W--I & 0.34 [0.27, 0.39] & 0.52 [0.47, 0.57] & yes (2/8) & 0.28 [0.23, 0.33] & yes (3/8) \\
W--F & 0.42 [0.32, 0.48] & $-$0.05 [$-$0.10, 0.00] & no, stronger in ratings (7/8) & 0.16 [0.12, 0.21] & no, stronger in ratings (4/8) \\
W--Imp & 0.10 [0.04, 0.17] & $-$0.25 [$-$0.30, $-$0.19] & no, opposite sign (3/8) & $-$0.26 [$-$0.32, $-$0.19] & no, opposite sign (2/8) \\
I--F & 0.44 [0.39, 0.49] & $-$0.02 [$-$0.07, 0.03] & no, stronger in ratings (7/8) & 0.24 [0.19, 0.29] & no, stronger in ratings (4/8) \\
I--Imp & 0.11 [0.05, 0.17] & $-$0.53 [$-$0.56, $-$0.49] & no, opposite sign (3/8) & $-$0.17 [$-$0.23, $-$0.11] & no, opposite sign (3/8) \\
F--Imp & $-$0.03 [$-$0.08, 0.03] & $-$0.27 [$-$0.32, $-$0.22] & yes (2/8) & $-$0.14 [$-$0.19, $-$0.09] & yes (3/8) \\
\bottomrule
\end{tabular}
\caption{Correlation between the ratings of two concepts in the same turn and between the corresponding probes at Prior and at Post (turn-adjusted Pearson, mean of the eight models, 95\% intervals from the bootstrap over topics), with ratings and probes oriented towards the high pole of each question. ``Explained'' is \emph{no} when the two intervals exclude 0 with opposite signs, or when the interval of $|r_{\text{ratings}}|-|r_{\text{probes}}|$ lies above 0; in parentheses, the number of models for which the same holds with each model's own interval. W, \emph{wellbeing}; I, \emph{interest}; F, \emph{focus}; Imp, \emph{impulsivity}.}\label{tab:pairs}
\end{table*}

\section{These limitations persist across model sizes}\label{app:scale}

\begin{figure*}[tp]
\centering\includegraphics[width=0.9\textwidth]{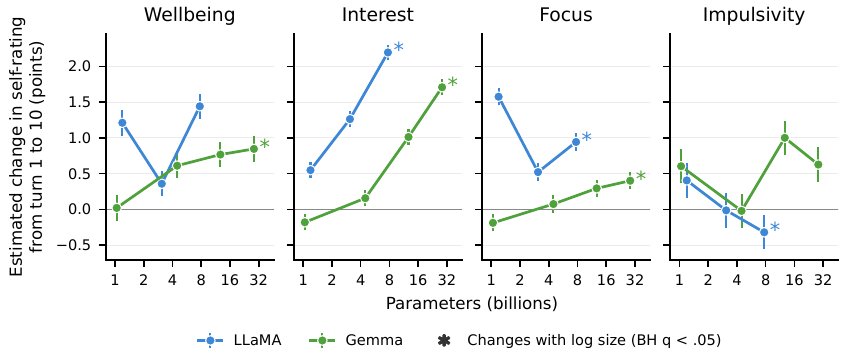}
\caption{Estimated change of the rating from turn 1 to turn 10 against the number of parameters, by concept.}\label{fig:scale_rating}
\end{figure*}

\begin{figure*}[tp]
\centering\includegraphics[width=0.9\textwidth]{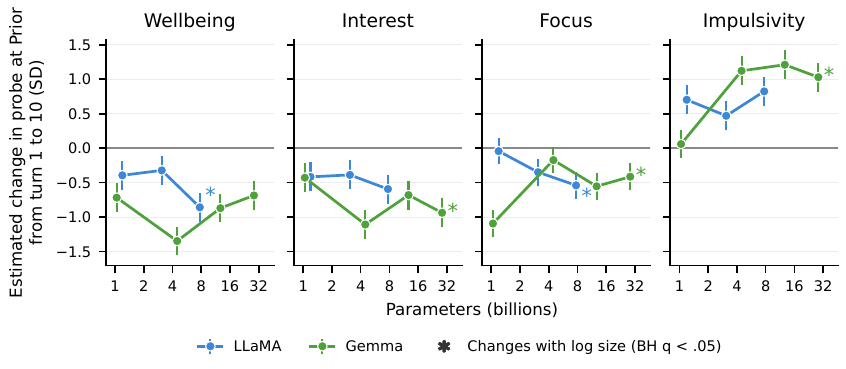}
\caption{Estimated change of the probe at Prior from turn 1 to turn 10 against the number of parameters, by concept.}\label{fig:scale_probe}
\end{figure*}

\begin{figure*}[tp]
\centering\includegraphics[width=0.9\textwidth]{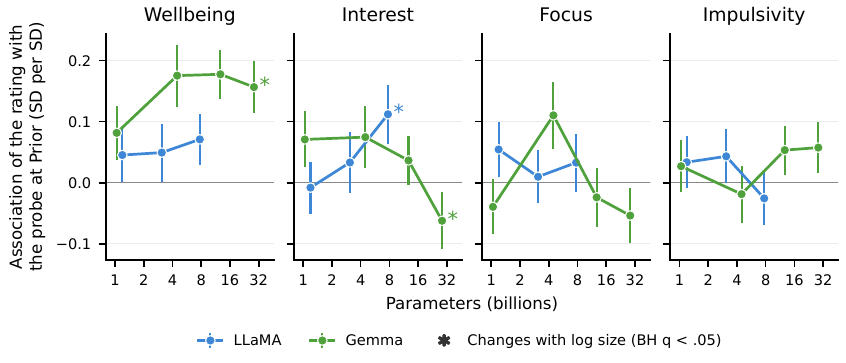}
\caption{Association of the rating with the probe at Prior (SD per SD, turn-adjusted) against the number of parameters, by concept.}\label{fig:scale_prior}
\end{figure*}

\begin{figure*}[tp]
\centering\includegraphics[width=0.9\textwidth]{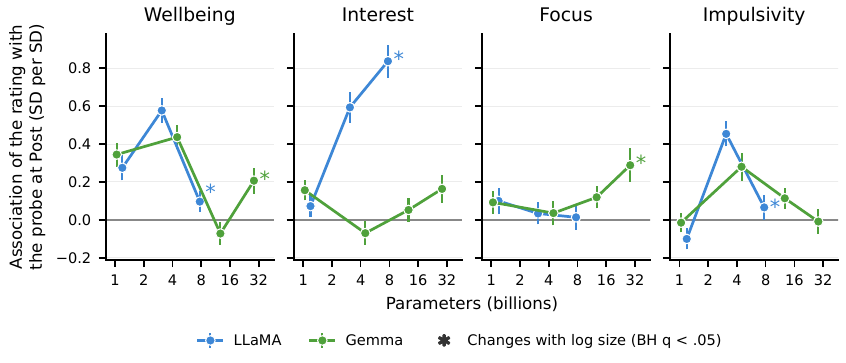}
\caption{Association of the rating with the probe at Post against the number of parameters, by concept.}\label{fig:scale_post}
\end{figure*}

\begin{figure*}[tp]
\centering\includegraphics[width=0.9\textwidth]{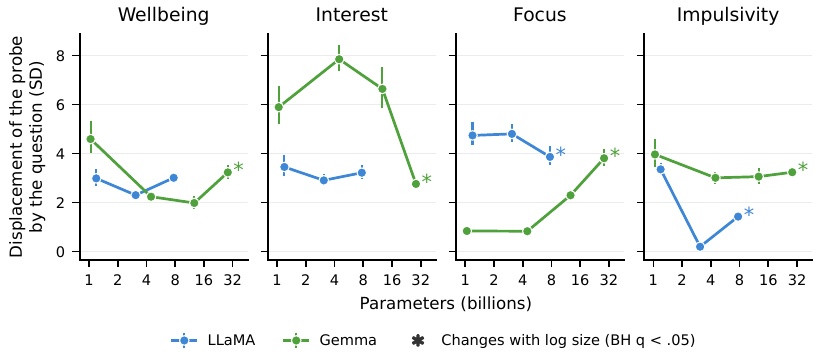}
\caption{Magnitude of the displacement of the probe at Post from Next after the affective question against the number of parameters, by concept.}\label{fig:scale_shift}
\end{figure*}

\begin{figure*}[tp]
\centering\includegraphics[width=0.9\textwidth]{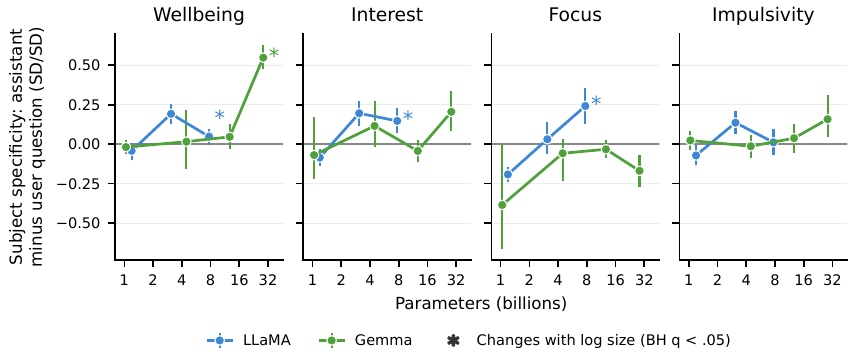}
\caption{Subject specificity against the number of parameters, by concept.}\label{fig:scale_subject}
\end{figure*}

\begin{figure*}[tp]
\centering\includegraphics[width=0.9\textwidth]{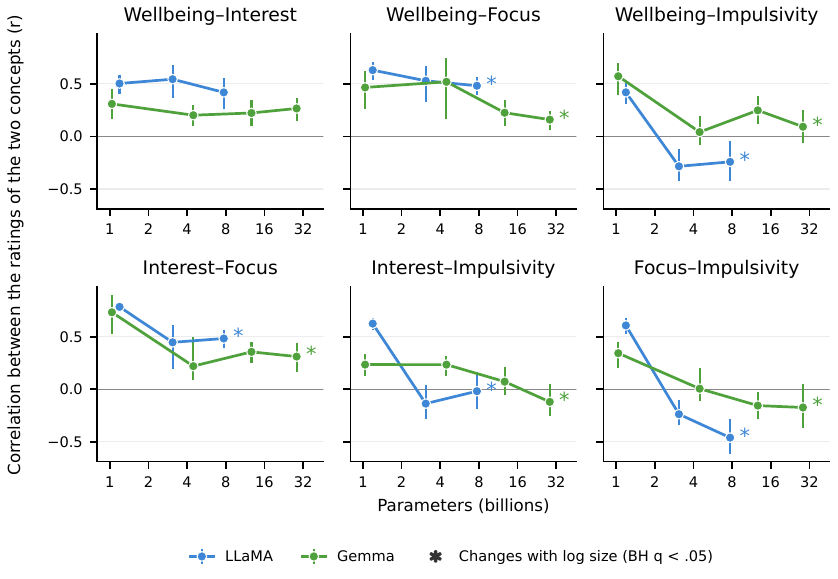}
\caption{Correlation between the ratings of two concepts in the same turn against the number of parameters, by pair.}\label{fig:scale_common}
\end{figure*}

\begin{figure*}[tp]
\centering\includegraphics[width=0.9\textwidth]{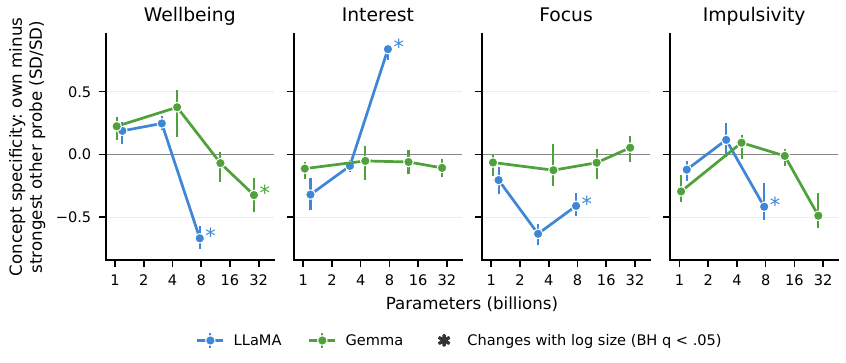}
\caption{Concept specificity against the number of parameters, by concept.}\label{fig:scale_concept}
\end{figure*}

Figure~\ref{fig:4} averages each quantity over the four concepts, so we asked whether its trends hold concept by concept. Table~\ref{tab:scale} gives the slopes behind Fig.~\ref{fig:4}, and Figs.~\ref{fig:scale_rating}--\ref{fig:scale_concept} break each quantity down by concept (by pair for the correlation between ratings), with an asterisk in the colour of the family when the slope differs from zero (BH over the tests of each figure).

\begin{table*}[tp]
\centering\scriptsize
\setlength{\tabcolsep}{3.5pt}
\begin{tabular}{lcccc}
\toprule
Quantity & LLaMA & $q$ & Gemma & $q$ \\
\midrule
Rating drift magnitude (points) & \textbf{0.11} [0.05, 0.16] & $<.001$ & \textbf{0.15} [0.12, 0.17] & $<.001$ \\
Probe drift magnitude (SD) & \textbf{0.12} [0.06, 0.16] & $<.001$ & \textbf{0.04} [0.01, 0.06] & .018 \\
Association with Prior (SD/SD) & 0.010 [$-$0.002, 0.021] & .111 & $-$0.002 [$-$0.008, 0.004] & .466 \\
Association with Post (SD/SD) & \textbf{0.062} [0.045, 0.080] & $<.001$ & $-$0.001 [$-$0.010, 0.008] & .796 \\
Displacement magnitude (SD) & \textbf{$-$0.28} [$-$0.39, $-$0.19] & .001 & \textbf{$-$0.11} [$-$0.22, $-$0.01] & .041 \\
Subject specificity (SD/SD) & \textbf{0.08} [0.05, 0.10] & .002 & \textbf{0.05} [0.02, 0.07] & .002 \\
Correlation between ratings ($r$) & \textbf{$-$0.18} [$-$0.21, $-$0.14] & .001 & \textbf{$-$0.07} [$-$0.09, $-$0.04] & .001 \\
Concept specificity (SD/SD) & $-$0.02 [$-$0.04, 0.01] & .320 & \textbf{$-$0.03} [$-$0.04, $-$0.02] & .002 \\
\bottomrule
\end{tabular}
\caption{Change per doubling of the parameter count within each family for the eight quantities of Fig.~\ref{fig:4}: least-squares slope of the per-model values against $\log_2$(size), with its 95\% interval, and $q$ after BH over the 16 tests; bold if $q<.05$.}\label{tab:scale}
\end{table*}

\paragraph{The drift grows with size in several concepts, but not uniformly.}
The rating rises more over the conversation in larger models for \emph{interest} in both families (0.61 points per doubling in LLaMA and 0.40 in Gemma) and for \emph{wellbeing} and \emph{focus} in Gemma, whereas in LLaMA the rise of \emph{focus} becomes smaller with size and the rating of \emph{impulsivity} goes from rising at 1B to falling at 8B (all $q<.001$; Fig.~\ref{fig:scale_rating}). The probe score drifts more in larger models for \emph{wellbeing} and \emph{focus} in LLaMA and for \emph{interest} and \emph{impulsivity} in Gemma, and less for \emph{focus} in Gemma (all $q\le.016$; Fig.~\ref{fig:scale_probe}). Some of these curves are not monotonic: the rating drift of LLaMA is smallest at 3B, and the probe drift of Gemma is largest at 4B. The larger drift of larger models in Fig.~\ref{fig:4}A and B therefore comes from several concepts, but not from all of them in both families.

\paragraph{No concept gains a substantial association with the state before the question.}
The association at Prior changes with size in three of the eight tests, by at most 0.04 SD/SD per doubling, and in opposite directions for \emph{interest}, which increases in LLaMA and decreases in Gemma (all $q\le.022$; Fig.~\ref{fig:scale_prior}). In larger models, the rating thus remains as weakly associated with the state before the question as in smaller ones.

\paragraph{The increase of the association at Post in LLaMA comes from two concepts.}
In LLaMA, the association at Post increases with size for \emph{interest} (0.29 per doubling) and \emph{impulsivity} (0.08) and decreases for \emph{wellbeing} ($-0.08$); in Gemma, it decreases for \emph{wellbeing} ($-0.06$) and increases slightly for \emph{focus} (0.03; all $q\le.008$; Fig.~\ref{fig:scale_post}). The mean of LLaMA also peaks at 3B (Fig.~\ref{fig:4}D), so its slope rests mainly on the difference between 1B and 8B.

\paragraph{The displacement shrinks with size in most concepts.}
The magnitude of the displacement decreases with size for \emph{focus} and \emph{impulsivity} in LLaMA and for \emph{wellbeing}, \emph{interest} and \emph{impulsivity} in Gemma, and increases for \emph{focus} in Gemma (all $q\le.008$; Fig.~\ref{fig:scale_shift}). The smaller displacement of larger models in Fig.~\ref{fig:4}E therefore holds in most, but not all, concepts.

\paragraph{Subject specificity increases with size in several concepts.}
In larger models, the gap between the association of the self-report with the rating of the assistant and its association with the rating of the user widens for \emph{wellbeing}, \emph{interest} and \emph{focus} in LLaMA and for \emph{wellbeing} in Gemma (0.03 to 0.16 SD/SD per doubling, all $q\le.012$; Fig.~\ref{fig:scale_subject}), and it narrows in no concept. The increase of Fig.~\ref{fig:4}F is therefore spread over several concepts.

\paragraph{Larger models give less correlated ratings across concepts.}
The correlation between the ratings of two concepts decreases with size in five of the six pairs in both families, all except \emph{wellbeing}--\emph{interest} (all $q\le.028$; Fig.~\ref{fig:scale_common}). In LLaMA, the decrease is largest in the pairs that involve \emph{impulsivity} ($-0.24$ to $-0.40$ per doubling, against $-0.03$ to $-0.11$ in the other pairs), and at 8B the correlation of the \emph{impulsivity} rating with the ratings of \emph{wellbeing}, \emph{interest} and \emph{focus} is $-0.24$, $-0.02$ and $-0.46$; in Gemma, the decreases are smaller ($-0.07$ to $-0.11$ in the five pairs). The component that the ratings share therefore weakens with size, most of all for \emph{impulsivity} in LLaMA.

\paragraph{Concept specificity changes with size in opposite directions across concepts.}
In LLaMA, concept specificity increases with size for \emph{interest} ($-0.32$, $-0.09$ and 0.84 for 1B, 3B and 8B) and decreases for the other three concepts, and in Gemma it decreases for \emph{wellbeing} (from 0.22 at 1B to $-0.32$ at 27B; all $q\le.013$; Fig.~\ref{fig:scale_concept}). The averages of Fig.~\ref{fig:4}H therefore combine opposite trends, and in neither family does the rating become specific to its own probe in all concepts.

\end{document}